\documentclass[letterpaper]{article} % DO NOT CHANGE THIS
\usepackage{aaai24}  % DO NOT CHANGE THIS
\usepackage{times}  % DO NOT CHANGE THIS
\usepackage{helvet}  % DO NOT CHANGE THIS
\usepackage{courier}  % DO NOT CHANGE THIS
\usepackage[hyphens]{url}  % DO NOT CHANGE THIS
\usepackage{graphicx} % DO NOT CHANGE THIS
\usepackage{natbib}  % DO NOT CHANGE THIS AND DO NOT ADD ANY OPTIONS TO IT
\usepackage{caption} % DO NOT CHANGE THIS AND DO NOT ADD ANY OPTIONS TO IT
\usepackage{algorithm}
\usepackage{algorithmic}

\usepackage{newfloat}
\usepackage{listings}
\DeclareCaptionStyle{ruled}{labelfont=normalfont,labelsep=colon,strut=off} % DO NOT CHANGE THIS
\floatstyle{ruled}
\newfloat{listing}{tb}{lst}{}
\floatname{listing}{Listing}
\usepackage{subfiles}
\usepackage{booktabs}       % professional-quality tables
\usepackage{amsmath,mathtools}
\usepackage{amsfonts}
\newcommand{\AlgComment}[1]{\hfill$\triangleright$ \textit{#1}}
\DeclareMathOperator*{\argmin}{argmin}
\usepackage{multirow}
\usepackage{xcolor}
\usepackage{adjustbox}

\title{SimPSI: A Simple Strategy to Preserve Spectral Information\\in Time Series Data Augmentation}
\author{
    Hyun Ryu,
    Sunjae Yoon,
    Hee Suk Yoon,
    Eunseop Yoon,
    and Chang D. Yoo
}
\affiliations{
    Korea Advanced Institute of Science and Technology (KAIST), Daejeon, South Korea
    \{ryuhyun1905, sunjae.yoon, hskyoon, esyoon97, cd\_yoo\}@kaist.ac.kr
}

\usepackage{bibentry}
\begin{document}

\maketitle

\begin{abstract}
Data augmentation is a crucial component in training neural networks to overcome the limitation imposed by data size, and several techniques have been studied for time series. Although these techniques are effective in certain tasks, they have yet to be generalized to time series benchmarks. We find that current data augmentation techniques ruin the core information contained within the frequency domain. To address this issue, we propose a simple strategy to preserve spectral information (\texttt{SimPSI}) in time series data augmentation. \texttt{SimPSI} preserves the spectral information by mixing the original and augmented input spectrum weighted by a preservation map, which indicates the importance score of each frequency. Specifically, our experimental contributions are to build three distinct preservation maps: magnitude spectrum, saliency map, and spectrum-preservative map. We apply \texttt{SimPSI} to various time series data augmentations and evaluate its effectiveness across a wide range of time series benchmarks. Our experimental results support that \texttt{SimPSI} considerably enhances the performance of time series data augmentations by preserving core spectral information. The source code used in the paper is available at \url{https://github.com/Hyun-Ryu/simpsi}.
\end{abstract}

\section{Introduction}

Time series data, whether univariate or multivariate, plays a crucial role in various domains such as medicine \cite{intro_medicine}, physiology \cite{intro_physiology}, and sensory devices \cite{intro_sensory_device}. Unfortunately, it is limited to collecting data samples under consideration of different types, constraining the performance and capabilities of neural networks that learn from it. To address this issue, data augmentation \cite{survey, um2017data} is employed as a simple yet effective solution via artificially increasing the number of samples based on a slight variation or perturbation on the original samples.

Data augmentation techniques have been extensively studied for time series, incorporating methods such as Jittering, Scaling, Magnitude warping, Time warping, Permutation \cite{um2017data}, Shifting \cite{cost}, and Dropout \cite{btsf}. These perturbations have been popular choices in the time domain. The data augmentation is also considered in the frequency domain via applying the Fourier transform to time series data. The spectrum is then randomly perturbed before being converted back into the time domain through the inverse Fourier transform. Notable techniques in this category include Frequency masking, Frequency mixing \cite{fraug}, and Frequency adding \cite{tfc}.

\begin{figure}[t]
  \centering
  \includegraphics[width=\columnwidth]{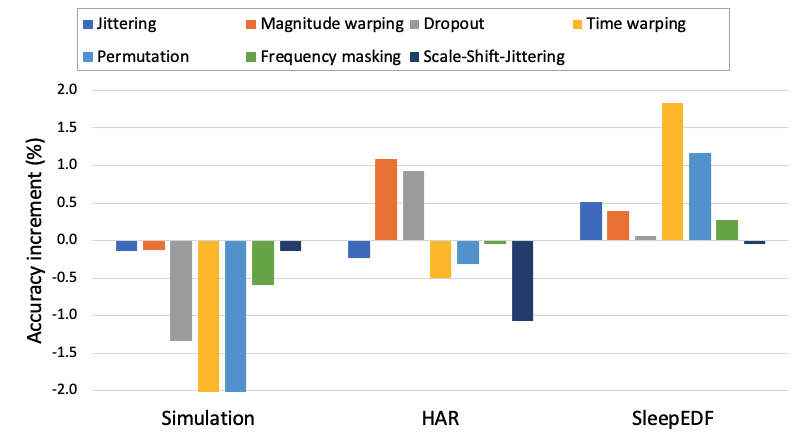}
  \caption{
  Dependency on data domain of time series data augmentation techniques. The plot shows the increment of classification accuracy of a baseline model after applying each data augmentation technique, which is evaluated on signal demodulation (Simulation), human activity recognition (HAR), and sleep stage detection (SleepEDF) tasks.
  }
  \label{fig:augs_pitfall}
\end{figure}

We have discovered that while the aforementioned data augmentation techniques show effectiveness in certain specific tasks \cite{um2017data}, they do not generalize well to time series classification benchmarks. Our experimental evidence in Fig. \ref{fig:augs_pitfall} presents the ungeneralized effectiveness of data augmentation techniques according to the datasets, such as signal demodulation, human activity recognition, and sleep stage detection.\footnote{Detailed information about the tasks and our experimental setup can be found in the Experiments section.} Those techniques, though reliant on randomness, operate under the assumption that the core information within the data is preserved. However, the result suggests that perturbing the original time series data is heuristic and depends on the data domain, which leads to losing essential information necessary to solve the tasks.

\begin{figure}[t]
  \centering
  \includegraphics[width=0.8\columnwidth]{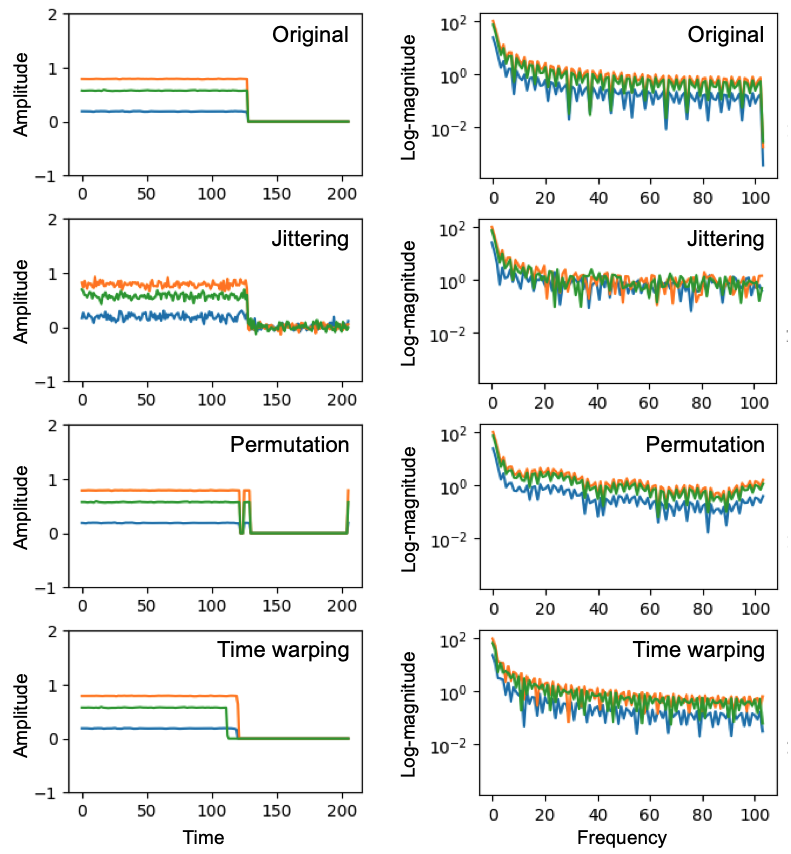}
  \caption{
  Visualization of a representative example from the HAR dataset in the time and frequency domain with various time series data augmentation techniques. Each color denotes a channel, and three channels are shown.
  }
  \label{fig:existing_augs}
\end{figure}

The observed reduction in performance is attributed to an implicit bias in the frequency domain introduced by each data augmentation technique. This bias alters the original data distribution. For example, Jittering adds a consistent amount of random noise across all frequencies, often obscuring subtle high-frequency components. Permutation, meanwhile, introduces abrupt changes at the boundaries of each fragment, consistently enhancing high-frequency components. Time warping globally distorts the temporal density of the original data, introducing even more spectral bias than Permutation. Fig. \ref{fig:existing_augs} provides illustrative examples of these data augmentation techniques.

In this paper, we introduce a simple strategy for preserving spectral information during time series data augmentation, which we refer to as \texttt{SimPSI}. Our strategy involves mixing the original spectral data and its augmented form, weighted by a preservation map.
After applying any time series data augmentation technique, \texttt{SimPSI} converts the original and augmented time series to the frequency domain. It then combines the original spectrum with the augmented version based on the weightage given by the preservation map, which indicates the importance score for each frequency component. The combined spectrum is subsequently transformed back to the time domain, resulting in the final output of our framework.
The remaining efforts concentrate on defining a well-structured preservation map. We propose three types of preservation maps: magnitude spectrum, saliency map, and spectrum-preservative map. The first two types use the given data's magnitude spectrum and saliency map \cite{saliency} as the preservation map. For the spectrum-preservative map, we developed a preservation map generator that takes input spectrum data and returns the preservation map. This map is learned through a preservation contrastive loss function that influences differentiated model output scores based on the preservation quality. We also propose a training strategy for improved optimization.
To demonstrate the efficacy of \texttt{SimPSI}, we apply it to various time series data augmentation techniques and compare performance across different benchmarks. We also create a simulation to assess whether the proposed method correctly identifies spectral regions to preserve during data augmentation. Our experimental results demonstrate that \texttt{SimPSI} significantly enhances the effectiveness of time series data augmentation techniques by preserving essential spectral information, thereby preventing unintentional loss of core spectral details.

\section{Related Works}

\begin{figure*}[t]
  \centering
  \includegraphics[width=\textwidth]{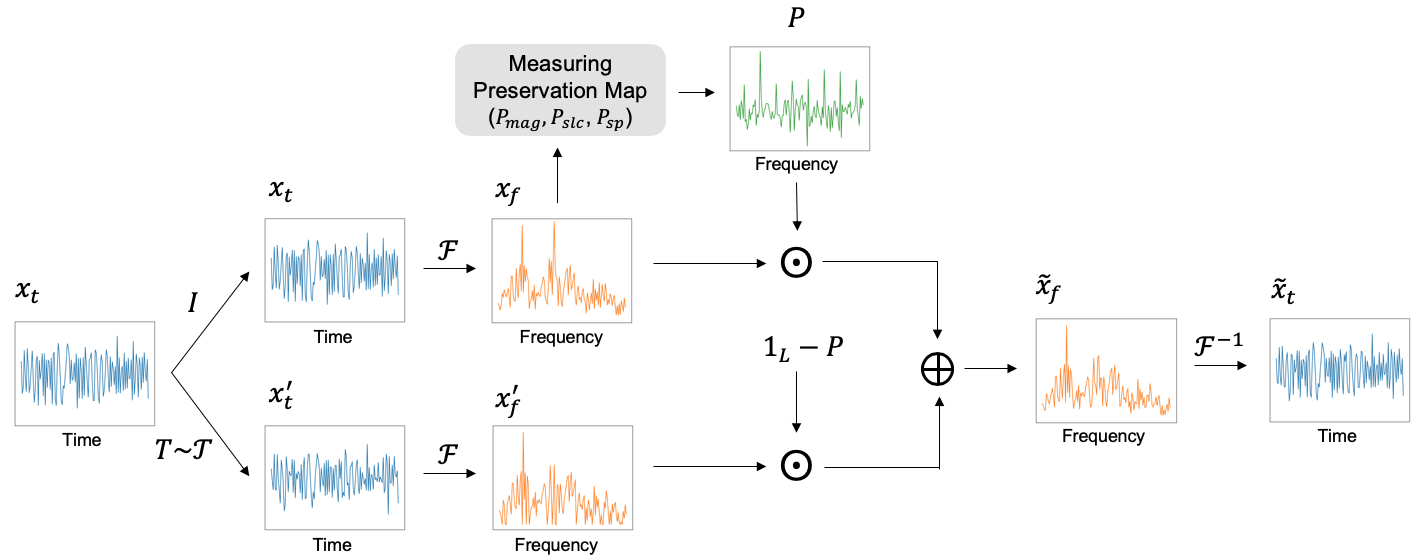}
  \caption{
  A \texttt{SimPSI} diagram. The original data is augmented randomly in the time domain. Then, the original and augmented data are both Fourier-transformed. The original spectrum is weighted by its preservation map, while the augmented spectrum is weighted by the negated preservation map, and those two are added. It is inverse-Fourier-transformed, which generates an information-preserved augmented view of the original time series data. We use a single-channel time series for better understanding, in which we visualize the real parts of the time series and magnitudes of spectra and omit channel-wise broadcasting.
  }
  \label{fig:simpsi}
\end{figure*}

\subsection{Data Augmentation for Time Series}
Various data augmentation techniques have been proposed for time series. One prevalent and intuitive strategy involves slightly altering the magnitude. For instance, Jittering \cite{um2017data} introduces additive white Gaussian noise, Scaling \cite{um2017data} multiplies by a random scalar value, Shifting \cite{cost} adds a random scalar value, Magnitude warping \cite{um2017data} multiplies by a random polynomial curve, and Dropout \cite{btsf} masks random time indices.
An alternative approach involves modifying the time scale rather than the magnitude. Time warping \cite{um2017data}, for instance, interpolates the time scale with a random polynomial curve, while Permutation \cite{um2017data} rearranges the time order.
An additional method involves perturbing the spectrum. Techniques such as Frequency masking, Frequency mixing \cite{fraug}, and Frequency adding \cite{tfc} serve as simple strategies that appropriately perturb global dependencies in the time domain.

\subsection{Data Augmentation for Information Preservation}
Data augmentation inherently introduces perturbations into the original data. If not appropriately managed, these perturbations could lead to significant information loss or an introduction of unnecessary noise and ambiguity. To mitigate this, studies have focused on information preservation.
In the vision domain, KeepAugment \cite{keepaugment} employs a saliency map \cite{saliency} of each image to identify and preserve informative regions during augmentation.
AugMix \cite{augmix} generates a composite of various augmented views of the data and mixes it with the original data, weighted by a random scalar. This ensures the final image is not overly distanced from the original one. In natural language processing, SSMix \cite{ssmix} and SMSMix \cite{smsmix} leverage saliency map to retain certain word sequences, ensuring that crucial information remains intact during the data augmentation process.
For time series data, Input smoothing \cite{stgcl} scales high-frequency entries in the frequency domain by a random scalar, thereby reducing the impact of data noise. However, its application is limited to noise reduction, and the degree of reduction is randomly determined.

\section{Method}

\subsection{Mixing for Information Preservation}

We transform an input time series $x_t \in \mathbb{C}^{C \times L}$, where $C$ and $L$ denote the number of channels and length of the input, to a spectrum $x_f \in \mathbb{C}^{C \times L}$ by the fast Fourier transform (FFT). Then, we apply data augmentation to $x_t$, which gives an augmented time series $x^{\prime}_t \in \mathbb{C}^{C \times L}$. We transform $x^{\prime}_t$ to an augmented spectrum $x^{\prime}_f \in \mathbb{C}^{C \times L}$ by the FFT.
Then, we define a preservation map $P \in \mathbb{R}^{L}$ with the same length as the spectrum $x_f$, which indicates the importance score of each frequency component between 0 and 1.
We mix the spectrum $x_f$ and its augmented view $x^{\prime}_f$ with the preservation map $P$ to produce an information-preserved spectrum $\tilde{x}_f \in \mathbb{C}^{C \times L}$ as follows:
\begin{gather}
\label{eq:mix}
\tilde{x}_f = (\mathbf{1}_{C} \cdot P^T) \odot x_{f} + (\mathbf{1}_{C} \cdot (\mathbf{1}_{L}-P)^T) \odot x^{\prime}_f.
\end{gather}
Since the preservation map $P$ applies uniformly to different channels of the spectra, we broadcast it to the channel dimension to enable elementwise multiplication with the spectra. Frequencies with high importance score have a spectrum value closer to the data $x_f$ than its augmentation $x^{\prime}_f$, and those with low importance score have a spectrum value closer to $x^{\prime}_f$ than $x_f$. It enables us to retain important spectral regions and distort non-informative regions during augmentation. We transform $\tilde{x}_f$ back to an information-preserved time series $\tilde{x}_t \in \mathbb{C}^{C \times L}$ by applying inverse fast Fourier transform (IFFT), which is the final output of the proposed \texttt{SimPSI}.
For classifier training, given a classifier $\hat{p}$, classification loss $\mathcal{L}_{cl}$ is calculated using the cross-entropy loss of the prediction score of $\tilde{x}_t$ and its label $y$ as follows:
\begin{gather}
\label{eq:cl_loss}
\mathcal{L}_{cl} = \mathcal{L}_{ce}(\hat{p}(y|\tilde{x}_t), y).
\end{gather}
The following sections focus on defining the preservation map $P$, and we propose three methods: magnitude spectrum, saliency map, and spectrum-preservative map.

\subsubsection{Efficient Implementation for Real-Valued Time Series.}
Most of the real-world time series data consists of real values. Using the conjugate symmetry property of the Fourier transform for real-valued time series, we take the first half of the spectrum, in which the dimensions of the spectrum reduce to $\{x_f, x^{\prime}_f, \tilde{x}_f\} \in \mathbb{C}^{C \times (\lfloor L/2 \rfloor+1)}$ while the dimensions of time series change to $\{x_t, x^{\prime}_t, \tilde{x}_t\} \in \mathbb{R}^{C \times L}$. The dimension of the preservation map also reduces to $P \in \mathbb{R}^{\lfloor L/2 \rfloor+1}$.

\subsection{Magnitude Spectrum}
We introduce a magnitude spectrum $P_{mag}$ for preserving spectral information, assuming frequencies with large magnitudes are informative while those with small magnitudes are mainly non-informative noise.
Given an input spectrum $x_f \in \mathbb{C}^{C \times L}$, we calculate the magnitude spectrum $|x_f| \in \mathbb{R}^{C \times L}$ and take the channel-wise maximum $|x_f|_{max} \in \mathbb{R}^{L}$ to aggregate the channel information as follows:
\begin{gather}
\label{eq:magprior}
P_{mag} = Norm(|x_f|_{max})
\end{gather}
where $Norm$ is a min-max normalization so that values of the magnitude spectrum $P_{mag} \in \mathbb{R}^{L}$ are between 0 and 1.
Preserving frequencies with large magnitudes makes the original and augmented data look alike, but the core information for solving the task might disappear. For instance, detecting abnormalities in the Electrocardiogram (ECG) signals relies on capturing the pattern of small high-frequency components \cite{hfecg}, whereas the magnitude spectrum $P_{mag}$ eliminates the core frequencies for the classification during the data augmentation process just because those have a small magnitude.

\subsection{Saliency Map}
We present a saliency map for time series, $P_{slc}$, to find informative spectral regions regardless of their magnitudes. Given an input spectrum $x_f \in \mathbb{C}^{C \times L}$, we transform it to a time series $x_t \in \mathbb{C}^{C \times L}$ by the IFFT and feed $x_t$ into the classifier $\hat{p}$ to obtain the corresponding label logit value $\hat{f}(y|x_{t})$. Then, we calculate an absolute value of a gradient of the logit value $\hat{f}(y|x_{t})$ with respect to the input spectrum $x_f$ and take the channel-wise maximum $|\nabla_{x_{f}} \hat{f}(y|x_t)|_{max} \in \mathbb{R}^{L}$ to aggregate the channel information as follows:
\begin{gather}
\label{eq:salprior}
P_{slc} = Norm(|\nabla_{x_{f}} \hat{f}(y|\mathcal{F}^{-1}(x_{f}))|_{max})
\end{gather}
where $Norm$ is a min-max normalization to make values of the saliency map $P_{slc} \in \mathbb{R}^{L}$ between 0 and 1.
However, it has a practical problem that the preservation quality solely depends on the training dynamics of the classifier, which could lead to an unstable performance. In addition, calculating the saliency map takes a significant amount of time backpropagating the gradients, which incurs a computational burden.

\begin{algorithm}[t]
\caption{\texttt{SimPSI} (Spectrum-Preservative Map) Pseudocode}
\label{alg:main}
\begin{algorithmic}
\STATE {\bf Input}: Given an input time series $x_t$, label $y$, preservation map generator $G(\cdot)$, classifier $\hat{p}$, and data augmentation $\mathcal{T}$
\STATE {\bf function} AugmentAndPreserve($x_t$, $x_f$, $P$)
\STATE ~~~~~~Sample operation $T \sim \mathcal{T}$
\STATE ~~~~~~$x^{\prime}_{t} = T(x_t)$ \AlgComment{Apply data augmentation}
\STATE ~~~~~~$x^{\prime}_{f} =$ FFT$(x^{\prime}_{t})$
\STATE ~~~~~~$\tilde{x}_f = (\mathbf{1}_{C} \cdot P^T) \odot x_{f} + (\mathbf{1}_{C} \cdot (\mathbf{1}_{L}-P)^T) \odot x^{\prime}_f$
\STATE ~~~~~~$\tilde{x}_t =$ IFFT$(\tilde{x}_f)$
\STATE ~~~~~~{\bf return} $\tilde{x}_t$
\STATE {\bf end function}
\STATE $x_f =$ FFT$(x_t)$
\STATE $\tilde{x}_{t}$ = AugmentAndPreserve($x_t$, $x_f$, $G(x_{f})$)
\STATE Compute classification loss $\mathcal{L}_{cl} = \mathcal{L}_{ce}(\hat{p}(y|\tilde{x}_{t}), y)$
\STATE Sample random preservation map $n_f \sim U(0,1)$
\STATE $\tilde{x}^{rnd}_{t}$ = AugmentAndPreserve($x_t$, $x_f$, $n_f$)
\STATE $\tilde{x}^{+}_{t}$ = AugmentAndPreserve($x_t$, $x_f$, $G(x_{f})$) \AlgComment{$\tilde{x}_{t} \neq \tilde{x}^{+}_{t}$}
\STATE $\tilde{x}^{-}_{t}$ = AugmentAndPreserve($x_t$, $x_f$, $1-G(x_{f})$)
\STATE Compute classification loss $\mathcal{L}_{cl}^{rnd}$, $\mathcal{L}_{cl}^{+}$, and $\mathcal{L}_{cl}^{-}$
\STATE for $\tilde{x}^{rnd}_{t}$, $\tilde{x}^{+}_{t}$, and $\tilde{x}^{-}_{t}$, respectively
\STATE Compute preservation contrastive loss $\mathcal{L}_{pc}$
\STATE $= max(\mathcal{L}_{cl}^{+} - \mathcal{L}_{cl}^{rnd} + \beta_1, 0) + max(\mathcal{L}_{cl}^{+} - L_{cl}^{-} + \beta_2, 0)$
\STATE {\bf Loss output}: $\mathcal{L}_{cl}$, $\mathcal{L}_{pc}$
\end{algorithmic}
\end{algorithm}

\subsection{Spectrum-Preservative Map}
We introduce a spectrum-preservative map $P_{sp}$, incorporating a preservation map generator $G$ on top of the classifier $\hat{p}$, alleviating the unstable training dynamics of the saliency map. It is also a feedforward network that does not require any additional backpropagation of the classifier $\hat{p}$ during estimating the preservation map, resolving the computational burden. The following describes how to design the preservation map generator $G$, what objective functions are used, and how to train it with the classifier $\hat{p}$.

\subsubsection{Preservation Map Generator.}
Given an input spectrum $x_f \in \mathbb{C}^{C \times L}$, we concatenate real and imaginary parts of $x_f$ into a channel dimension, which the dimension changes to $x_f \in \mathbb{R}^{2C \times L}$. Then, $x_f$ is fed into a two-layer transformer encoder to capture the underlying context of spectral representation. The output of the last layer is averaged over the channel dimension to aggregate the channel information and passes through the sigmoid function to make the values between 0 and 1 as follows:
\begin{gather}
\label{eq:selfprior}
P_{sp} = G(x_{f}) = Sigmoid(Enc(x_{f})_{mean}).
\end{gather}

\subsubsection{Preservation Contrastive Loss.}
To train the preservation map generator $G$, we introduce a preservation contrastive loss $\mathcal{L}_{pc}$. Assume that an input spectrum $x_f \in \mathbb{C}^{C \times L}$, augmented spectrum $x^{\prime}_f \in \mathbb{C}^{C \times L}$, and corresponding spectrum-preservative map $P_{sp}$ are given.
We define an information-preserved spectrum $\tilde{x}^{+}_{f} = (\mathbf{1}_{C} \cdot P_{sp}^T) \odot x_{f} + (\mathbf{1}_{C} \cdot (\mathbf{1}_{L}-P_{sp})^T) \odot x^{\prime}_f$ and a spectrum that preserves the inverted preservation map $\tilde{x}^{-}_{f} = (\mathbf{1}_{C} \cdot (\mathbf{1}_{L}-P_{sp})^T) \odot x_{f} + (\mathbf{1}_{C} \cdot P_{sp}^T) \odot x^{\prime}_f$. Then, the classifier $\hat{p}$ should predict $\tilde{x}^{+}_{t}$ better than $\tilde{x}^{-}_{t}$.
Furthermore, if we define a randomly-preserved spectrum $\tilde{x}^{rnd}_{f} = (\mathbf{1}_{C} \cdot n_f^T) \odot x_{f} + (\mathbf{1}_{C} \cdot (\mathbf{1}_{L}-n_f)^T) \odot x^{\prime}_f$ where $n_f \in \mathbb{R}^{L}$ is a random noise sampled from $U(0,1)$, then the prediction score of $\tilde{x}^{rnd}_{t}$ by the classifier $\hat{p}$ should be in between those of $\tilde{x}^{+}_{t}$ and $\tilde{x}^{-}_{t}$. We notate $\tilde{x}^{\{+, -, rnd\}}_{t}$ and $\tilde{x}^{\{+, -, rnd\}}_{f}$ as Fourier transform pairs of time series and corresponding spectrum. These constraints can be formulated as follows:
\begin{equation}
\label{eq:ctr}
\hat{p}(y|\tilde{x}^{+}_{t}) > \hat{p}(y|\tilde{x}^{rnd}_{t}) > \hat{p}(y|\tilde{x}^{-}_{t}).
\end{equation}
We then define the corresponding classification loss using the cross-entropy loss for these three predictions, $\mathcal{L}_{cl}^{+} = \mathcal{L}_{ce}(\hat{p}(y|\tilde{x}^{+}_{t}), y)$, $\mathcal{L}_{cl}^{rnd} = \mathcal{L}_{ce}(\hat{p}(y|\tilde{x}^{rnd}_{t}), y)$, and $\mathcal{L}_{cl}^{-} = \mathcal{L}_{ce}(\hat{p}(y|\tilde{x}^{-}_{t}), y)$ where $y$ is the label of the input time series $x_t$, and translate the constraints into an objective function as follows:
\begin{multline}
\label{eq:ctrloss}
\mathcal{L}_{pc} = max(\mathcal{L}_{cl}^{+} - \mathcal{L}_{cl}^{rnd} + \beta_1, 0)\\
+ max(\mathcal{L}_{cl}^{+} - \mathcal{L}_{cl}^{-} + \beta_2, 0)
\end{multline}
where $\beta_1$ and $\beta_2$ are hyperparameters satisfying $\beta_1 < \beta_2$.

\subsubsection{Model Training and Inference.}
We use two objective functions for model training, classification loss $\mathcal{L}_{cl}$ for classifier training and the preservation contrastive loss $\mathcal{L}_{pc}$ for preservation map generator training.
We separate the training procedure, updating the classifier $\hat{p}$ by $\mathcal{L}_{cl}$ with the preservation map generator $G$ froze, and then updating the preservation map generator $G$ by $\mathcal{L}_{pc}$ with the classifier $\hat{p}$ froze. It can be formulated as follows:
\begin{equation}
\begin{aligned}
\hat{\theta}_{p} &= \argmin_{\theta_{p}} \mathcal{L}_{cl}(x_t|\theta_{G},\theta_{p})\\
\hat{\theta}_{G} &= \argmin_{\theta_{G}} \mathcal{L}_{pc}(x_t|\theta_{G},\theta_{p})
\end{aligned}
\end{equation}
where $\theta_{G}$ and $\theta_{p}$ are the parameters of the preservation map generator $G$ and the classifier $\hat{p}$, respectively.
Note that $\theta_{G}$ does not descend towards the gradient of $\mathcal{L}_{cl}$. This prevents $G$ from learning undesirable local minima, such as returning the uniform scalar value across different frequencies or the same map across different samples, in which the preservation map is not adaptive to the input time series but acts as a uniform band-pass filter. Also, the preservation map generator $G$ is removed during inference, so $G$ updated by the classification loss might interrupt classifier training.

\section{Experiments}

\subsection{Signal Demodulation (Simulation)}

\begin{figure}[t]
  \centering
  \includegraphics[width=\columnwidth]{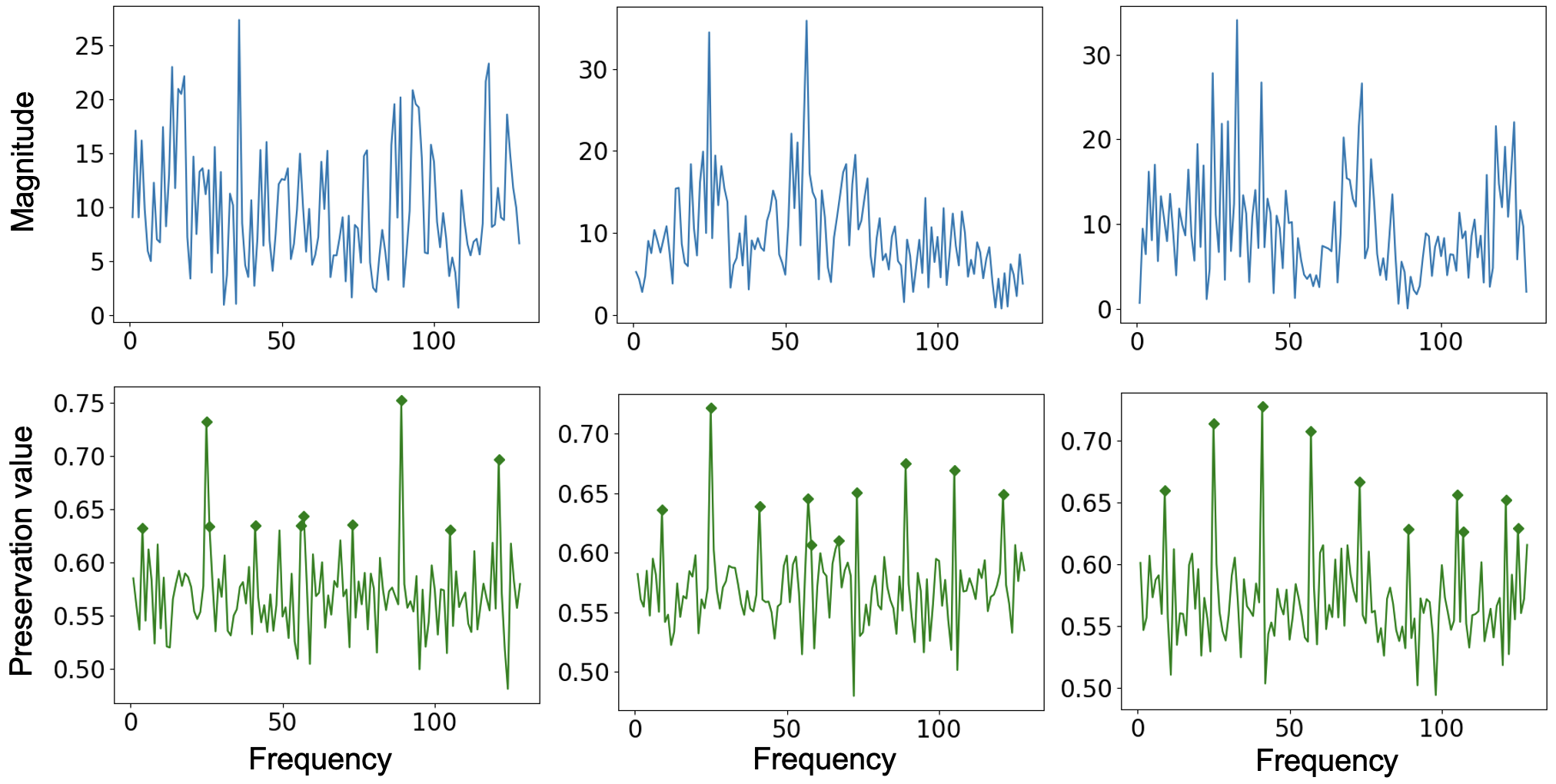}
  \caption{
  Finding a set of frequencies to preserve using \texttt{SimPSI} (Spectrum-preservative map) during Frequency masking. The top row shows representative input magnitude spectra from the FSK8 test set. The bottom row shows the corresponding learned preservation map where the ten largest values are marked as diamonds.
  }
  \label{fig:toy}
\end{figure}

\subsubsection{Experimental Setting.}
We verified if the proposed method improves the performance of existing data augmentation techniques by capturing important spectral regions and preserving them. To do that, we devised a simulation where information is carried on a set of known frequencies. Inspired by the wireless communication domain \cite{emc2net}, we constructed a synthetic dataset by modulating a sequence of random bits into the corresponding frequencies of a signal, called frequency shift keying (FSK), and the task is demodulating it.
We used 8 and 32 different frequencies for modulation (FSK8 and FSK32), and each dataset consists of 2,304 training signals, 288 validation signals, and 288 testing signals. We chose ResNet1D \cite{resnet1d} as a baseline network, which was given a 128-length modulated signal and returned a 32-length M-ary (M=8, 32) sequence. We used the Adam optimizer \cite{adam} with the learning rate $10^{-3}$, and the networks were trained for 50 epochs. The training was performed on a single NVIDIA RTX A6000 GPU. Appendix A provides more details about our experimental setup.

\subsubsection{Performance Enhancement through SimPSI.}
The performance improvement of random augmentations by the proposed method on the FSK32 dataset is described in Table \ref{table:datasets}. The accuracies of Jittering, Scale-Shift-Jittering, and Frequency masking were increased by 1.5\%, 1.4\%, and 1.5\%, respectively, using the spectrum-preservative map.

\subsubsection{Learned Preservation Map.}
We then verified whether the learned preservation map genuinely preserves the informative frequency components during augmentation. We displayed learned preservation maps of representative samples from the FSK8 test set in Fig. \ref{fig:toy}. We could observe eight equally-spaced frequencies that were preserved the most during Frequency masking. It perfectly matches the data generation process since we used those eight frequencies for signal modulation.
The other frequencies did not contain information and showed a preservation value of around 0.5, meaning those components barely attributed to achieving Eq. (\ref{eq:ctr}).

\subsection{Human Activity Recognition (HAR)}

\begin{table*}[t]
  \centering
  \begin{tabular}{lcccccc}
    \toprule
    \multirow{2}{*}{Model} & \multicolumn{2}{c}{Simulation}    & \multicolumn{2}{c}{HAR}       & \multicolumn{2}{c}{SleepEDF}\\
     & $Accuracy$ & $\Delta$ & $Accuracy$ & $\Delta$ & $Accuracy$ & $\Delta$ \\
    \midrule
    None            & 94.8 $\pm$ 0.1 & N/A & 94.0 $\pm$ 0.8 & N/A & 80.7 $\pm$ 0.1 & N/A \\
    \midrule
    Jittering \cite{um2017data}   & 94.6 $\pm$ 0.1 & \textminus0.2 & 93.8 $\pm$ 0.8 & \textminus0.2 & 81.2 $\pm$ 0.4 & +0.5 \\
    + Random preservation map                  & 96.2 $\pm$ 0.3 & +1.4 & 93.6 $\pm$ 0.5 & \textminus0.4 & 81.2 $\pm$ 0.5 & +0.5 \\
    + \texttt{SimPSI} (Magnitude spectrum)          & 94.9 $\pm$ 0.3 & +0.1 & \textbf{94.2 $\pm$ 0.5} & \textbf{+0.2} & 81.0 $\pm$ 0.3 & +0.3 \\
    + \texttt{SimPSI} (Saliency map)                & 96.2 $\pm$ 0.2 & +1.4 & 93.7 $\pm$ 0.6 & \textminus0.3 & \textbf{81.4 $\pm$ 0.4} & \textbf{+0.7} \\
    + \texttt{SimPSI} (Spectrum-preservative map)   & \textbf{96.3 $\pm$ 0.1} & \textbf{+1.5} & 94.1 $\pm$ 0.3 & +0.1 & 81.3 $\pm$ 0.4 & +0.6 \\
    \midrule
    Scale-Shift-Jittering \cite{cost} & 94.6 $\pm$ 0.2 & \textminus0.2 & 92.8 $\pm$ 0.5 & \textminus1.2 & 80.7 $\pm$ 0.2 & 0 \\
    + Random preservation map                & 95.3 $\pm$ 0.1 & +0.5 & 94.0 $\pm$ 1.3 & 0 & 80.4 $\pm$ 1.5 & \textminus0.3 \\
    + \texttt{SimPSI} (Magnitude spectrum)        & 94.9 $\pm$ 0.1 & +0.1 & \textbf{94.9 $\pm$ 1.0} & \textbf{+0.9} & \textbf{81.4 $\pm$ 0.7} & \textbf{+0.7} \\
    + \texttt{SimPSI} (Saliency map)              & 95.6 $\pm$ 0.3 & +0.8 & 94.8 $\pm$ 0.2 & +0.8 & 81.2 $\pm$ 0.5 & +0.5 \\
    + \texttt{SimPSI} (Spectrum-preservative map) & \textbf{96.2 $\pm$ 0.1} & \textbf{+1.4} & 94.8 $\pm$ 0.6 & +0.8 & 80.9 $\pm$ 1.0 & +0.2 \\
    \midrule
    Frequency masking \cite{fraug} & 94.2 $\pm$ 0.3 & \textminus0.6 & 93.9 $\pm$ 1.7 & \textminus0.1 & 81.0 $\pm$ 0.4 & +0.3 \\
    + Random preservation map                   & 96.2 $\pm$ 0.3 & +1.4 & \textbf{95.0 $\pm$ 0.6} & \textbf{+1.0} & 81.5 $\pm$ 0.5 & +0.8 \\
    + \texttt{SimPSI} (Magnitude spectrum)           & 94.9 $\pm$ 0.1 & +0.1 & \textbf{95.0 $\pm$ 0.4} & \textbf{+1.0} & 80.5 $\pm$ 0.4 & \textminus0.2 \\
    + \texttt{SimPSI} (Saliency map)                 & \textbf{96.3 $\pm$ 0.1} & \textbf{+1.5} & \textbf{95.0 $\pm$ 0.5} & \textbf{+1.0} & 80.2 $\pm$ 1.4 & \textminus0.5 \\
    + \texttt{SimPSI} (Spectrum-preservative map)    & \textbf{96.3 $\pm$ 0.2} & \textbf{+1.5} & \textbf{95.0 $\pm$ 0.5} & \textbf{+1.0} & \textbf{81.7 $\pm$ 0.2} & \textbf{+1.0} \\
    \bottomrule
  \end{tabular}
  \caption{Performance on Signal Demodulation (Simulation test set), Human Activity Recognition (HAR test set), and Sleep Stage Detection (SleepEDF test set) using different random augmentations with and without \texttt{SimPSI}. Accuracy and its increment from not using augmentation are reported with three different seeds.}
  \label{table:datasets}
\end{table*}

\begin{table}[th]
  \centering
  \begin{adjustbox}{width=\columnwidth}
  \begin{tabular}{lccc}
    \toprule
    \multirow{2}{*}{Model} & 3-layer    & 2-layer       & 2-layer\\
    & CNN & LSTM & Transformer \\
    \midrule
    Jittering \cite{um2017data} & 95.1 & 92.2 & 94.2 \\
    + \texttt{SimPSI} ($P_{sp}$)  & \textbf{95.2} & \textbf{93.7} & \textbf{96.4} \\
    \bottomrule
  \end{tabular}
  \end{adjustbox}
  \caption{Performance on Human Activity Recognition using various model architectures with and without \texttt{SimPSI} (Spectrum-preservative map). AUPRC scores are averaged over three different seeds.}
  \label{table:arch}
\end{table}

\subsubsection{Experimental Setting.}
In the HAR dataset \cite{ucihar}, data is collected by the accelerometer and gyroscope of a smartphone mounted on a waist and sampled at 50 Hz, which aims to classify human activities. Following the data preprocessing in \cite{tstcc}, an input time series has a length of 128 and nine channels. The dataset consists of 7,352 training samples and 2,947 test samples labeled with six classes.
We chose a 3-layer CNN model for classification, which was used in \cite{strongbaseline, tstcc, tfc}, and additionally included a 2-layer LSTM model and a 2-layer Transformer model for further verifications. We used the Adam optimizer with the learning rate $10^{-3}$, and the networks were trained for 100 epochs. We adhered to configurations in \cite{tstcc}, and the training was performed on a single NVIDIA RTX A6000 GPU. Appendix B and C provide more details about our experimental setup.

\begin{figure}[t]
  \centering
  \includegraphics[width=0.8\columnwidth]{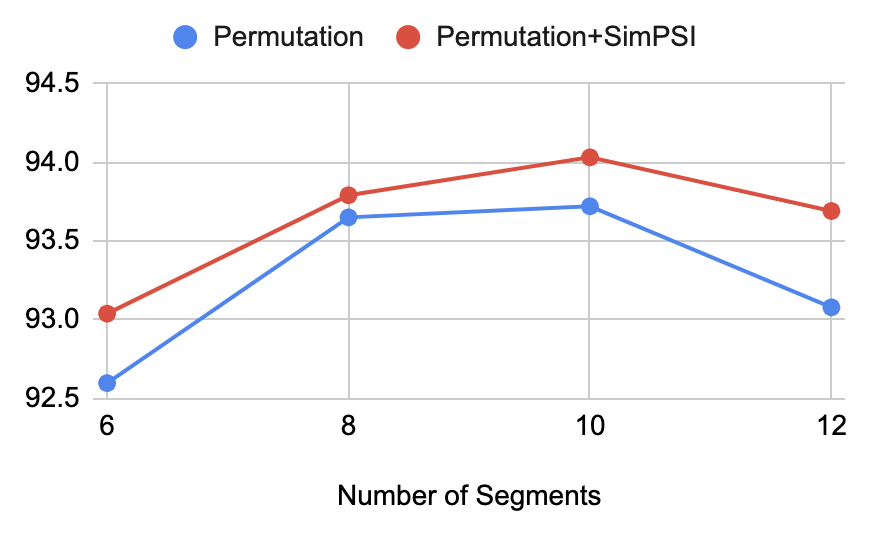}
  \caption{
  Testing accuracy of a 3-layer CNN model trained on the HAR dataset using Permutation with and without \texttt{SimPSI} (Spectrum-preservative map) while varying the maximum number of segments.
  }
  \label{fig:aug_strength}
\end{figure}

\subsubsection{Performance Enhancement through SimPSI.}
We compared the performance of the model with and without \texttt{SimPSI} to evaluate the impact of \texttt{SimPSI} on recognition accuracy. To inspect its impact thoroughly, we performed experiments on three perspectives: different random augmentations, model architectures, and distortion magnitudes.

The performance increase of data augmentations by \texttt{SimPSI} on the HAR dataset is described in Table \ref{table:datasets}. We also tested an intuitive method that mixes the original data and its augmented view with a random preservation map sampled from $U(0,1)$. The accuracy of Jittering is enhanced by 0.2\% using the magnitude spectrum, while the random preservation map decreases it by 0.4\%. The accuracy of Scale-Shift-Jittering is improved by 0.9\% using the magnitude spectrum, and Frequency masking is improved by 1.0\% using all the proposed preservation maps. Appendix D provides more experimental results, and Appendix E provides the training time cost of the preservation maps.

Using Jittering, we compared three types of networks for time series classification: CNN, LSTM, and Transformer (Table \ref{table:arch}). \texttt{SimPSI} increased the area under the precision-recall curve (AUPRC) of a 3-layer CNN by 0.1, while it increased the AUPRC of a 2-layer LSTM by 1.5 and a 2-layer Transformer by 2.2.

Using Permutation, we also tested \texttt{SimPSI} while varying the distortion magnitude of data augmentation. We changed the maximum number of segments of Permutation and compared the accuracy with and without \texttt{SimPSI} (Fig. \ref{fig:aug_strength}). \texttt{SimPSI} consistently improved the performance of Permutation regardless of its distortion strength, alleviating the performance drop while the number of segments increased. Specifically, comparing the performance at 10 and 12 segments, Permutation alone dropped the accuracy by 0.6, while Permutation with \texttt{SimPSI} dropped it by 0.3.

\subsection{Sleep Stage Detection (SleepEDF)}

\subsubsection{Experimental Setting.}
We used the SleepEDF dataset \cite{physiobank} for classifying sleep stages from Electroencephalogram (EEG) signals sampled at 100 Hz. We followed the data preprocessing in \cite{tstcc}, where the input has a length of 3,000 and a single channel. The dataset comprises 35,503 training samples and 6,805 test samples labeled with five classes.
We chose a 3-layer CNN model for classification, also used in the human activity recognition experiments. We used the Adam optimizer with the learning rate $10^{-3}$, and the networks were trained for 40 epochs. We adhered to configurations in \cite{tstcc} for other details. The training was performed on a single NVIDIA RTX A6000 GPU.

\subsubsection{Performance Enhancement through SimPSI.}
The performance improvement of data augmentations by \texttt{SimPSI} on the SleepEDF dataset is summarized in Table \ref{table:datasets}. \texttt{SimPSI} increased the detection accuracy of Jittering by 0.7\% using the saliency map, Scale-Shift-Jittering by 0.7\% using the magnitude spectrum, and Frequency masking by 1.0\% using the spectrum-preservative map. We note that the spectrum-preservative map outperformed the random preservation map regardless of the baseline augmentation techniques, supporting the effectiveness of the information-preserving approach.

\subsection{Atrial Fibrillation Classification (Waveform)}
We used the Waveform dataset \cite{waveform} for classifying rhythm types from ECG recordings of human subjects with atrial fibrillation. It was sampled at 250 Hz, and we followed the data preprocessing step as in \cite{tnc}. Every input has a length of 2,500 and two channels. The dataset comprises 59,922 training samples and 16,645 test samples labeled with four classes.
We chose a 1-dimensional strided CNN with six convolutional layers and a total down-sampling factor 16, proposed in \cite{tnc}. We used the Adam optimizer with the learning rate $10^{-4}$, and the networks were trained for 8 epochs. We adhered to configurations in \cite{tnc} for other details. The training was performed on a single NVIDIA RTX A6000 GPU.
Performance enhancement through \texttt{SimPSI} is described in Appendix D.

\begin{table}[t]
  \centering
  \begin{adjustbox}{width=\columnwidth}
  \begin{tabular}{lcc}
    \toprule
    Model & $Accuracy$ & $AUPRC$ \\
    \midrule
    Scale-Shift-Jittering                   & 92.0 $\pm$ 1.9 & 62.6 $\pm$ 2.3 \\
    + \texttt{SimPSI} ($P_{sp}$)                   & \textbf{95.2 $\pm$ 0.3} & \textbf{64.7 $\pm$ 2.0} \\
    + \texttt{SimPSI} ($P_{sp}$) w/o $\mathcal{L}_{pc}$      & 94.9 $\pm$ 0.1 & 63.7 $\pm$ 2.1 \\
    + \texttt{SimPSI} ($P_{sp}$) w/ joint training & 94.5 $\pm$ 0.3 & 62.9 $\pm$ 1.9 \\
    \bottomrule
  \end{tabular}
  \end{adjustbox}
  \caption{Ablation of \texttt{SimPSI} (Spectrum-preservative map) on Atrial Fibrillation Classification. Accuracy and AUPRC scores are reported with three different seeds.}
  \label{table:ablation}
\end{table}

\subsection{Ablations}
We ablated the proposed method from two perspectives, verifying the impact of the preservation contrastive loss and separate training strategy. We showed the performance of a 6-layer CNN model on the Waveform dataset while a composition of Scaling, Shifting, and Jittering \cite{cost} was applied (Table \ref{table:ablation}). Removing the preservation contrastive loss resulted in a 0.3 decrease in accuracy and a 1.0 decrease in AUPRC. Applying joint training of the cross-entropy loss and the preservation contrastive loss made a 0.7 decrease in accuracy and a 1.8 decrease in AUPRC.

\section{Discussions}

\begin{figure}[t]
  \centering
  \includegraphics[width=\columnwidth]{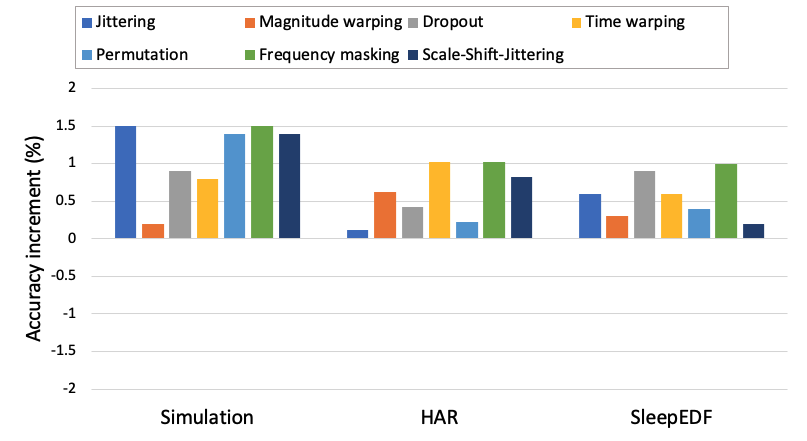}
  \caption{
  \texttt{SimPSI}'s dependency on data domain. The plot shows the increment of classification accuracy of a baseline model after applying each data augmentation technique with \texttt{SimPSI} (Spectrum-preservative map), which is evaluated on Simulation, HAR, and SleepEDF datasets.
  }
  \label{fig:simpsi_ddd}
\end{figure}

\begin{figure}[t]
  \centering
  \includegraphics[width=\columnwidth]{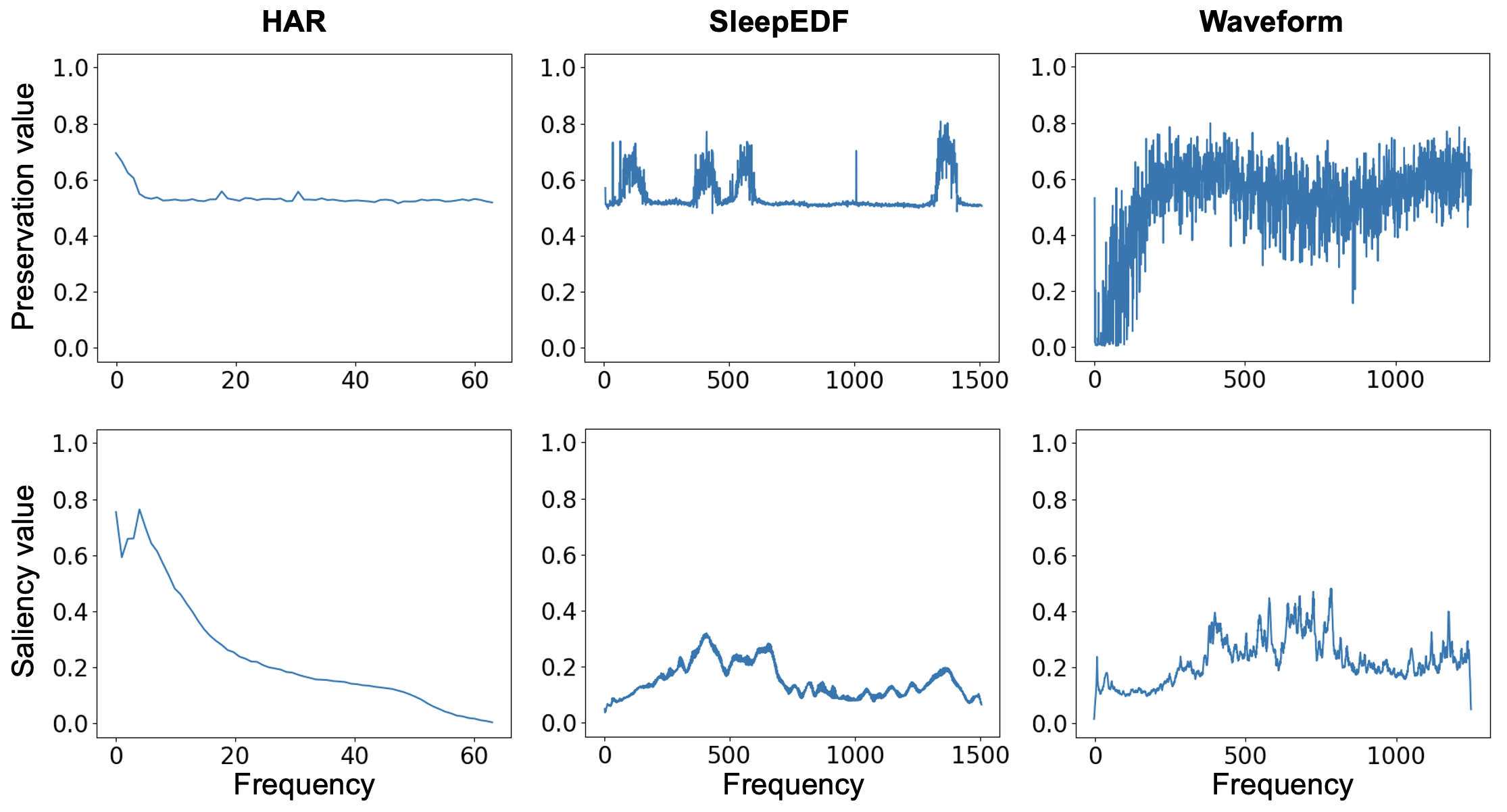}
  \caption{
  Comparison of spectrum-preservative maps and saliency maps. All the maps are averaged on the HAR, SleepEDF, and Waveform test sets. We used Jittering during training.
  }
  \label{fig:discussion}
\end{figure}

\subsubsection{SimPSI's Dependency on Data Domain.}
We observed the data augmentation techniques did not generalize well to time series benchmarks (Fig. \ref{fig:augs_pitfall}). Specifically, no augmentation increased the accuracy on the Simulation dataset. However, \texttt{SimPSI} resolved the issue in which the information-preserved approaches consistently improved the performance regardless of the tasks (Fig. \ref{fig:simpsi_ddd}). As a result, \texttt{SimPSI} encouraged augmentation independent of the data domain by preserving core spectral information. Appendix F provides more results using different preservation maps.

\subsubsection{Comparison of Preservation Maps.}
The averaged spectrum-preservative map for the HAR dataset showed that the few lowest frequencies were preserved better than the higher frequencies. The averaged saliency map showed a similar tendency, where the saliency value was high at the few lowest frequencies and fell to zero as the frequency increased.
For the SleepEDF dataset, there were four distinct frequency clusters in the spectrum-preservative map, and we could find corresponding clusters in the saliency map.
For the Waveform dataset, unlike the previous two datasets, high-frequency components are preserved more than the lower ones in both preservation maps. These are displayed in Fig. \ref{fig:discussion}.

\section{Conclusion}

We presented a simple strategy to preserve spectral information (\texttt{SimPSI}) in time series data augmentation.
Our investigation into the simulation task proved that the proposed method preserves the informative frequency components during augmentation. Our experimental results on various time series tasks with different data augmentation techniques illustrated the effectiveness of \texttt{SimPSI} in enhancing the model performance.
We believe that \texttt{SimPSI} is a powerful tool to mitigate the data domain dependency of time series data augmentation techniques and improve the model performance in various time series tasks.

\section{Acknowledgments}
This work was supported by Institute for Information \& communications Technology Planning \& Evaluation (IITP) grant funded by the Korea government (MSIT) (No. 2021-0-01381, Development of Causal AI through Video Understanding and Reinforcement Learning, and Its Applications to Real Environments) and partly supported by Institute of Information \& communications Technology Planning \& Evaluation (IITP) grant funded by the Korea government (MSIT) (No. 2022-0-00184, Development and Study of AI Technologies to Inexpensively Conform to Evolving Policy on Ethics).

\bibliography{aaai24}

% appendix
\clearpage
\appendix

\section{A. Details of Simulation Dataset}
This section gives a detailed explanation of constructing the Simulation dataset for signal demodulation.
We modulated the signal using frequency shift keying (FSK) to assign a sequence of bits to a sequence of frequencies of a signal. We used 8 and 32 different frequencies for signal modulation (i.e., FSK8, FSK32), where frequencies were separated by 16 Hz for FSK8 and 4 Hz for FSK32, while the sample rate was 128 Hz. The M-ary (M = 8, 32) sequence of random bits had a length of 32, and the samples per symbol rate was 4, which made the signal length 128. After signal modulation, we included an additive white Gaussian noise (AWGN) channel, in which the signal-to-noise ratio (SNR) varied from 10 to 28 dB. The signal was then normalized to unit power. Data is generated via MATLAB, and we adapted the instructions given by \cite{emc2net}.

\section{B. Data Augmentations}
\begin{figure*}[h]
  \centering
  \includegraphics[width=\textwidth]{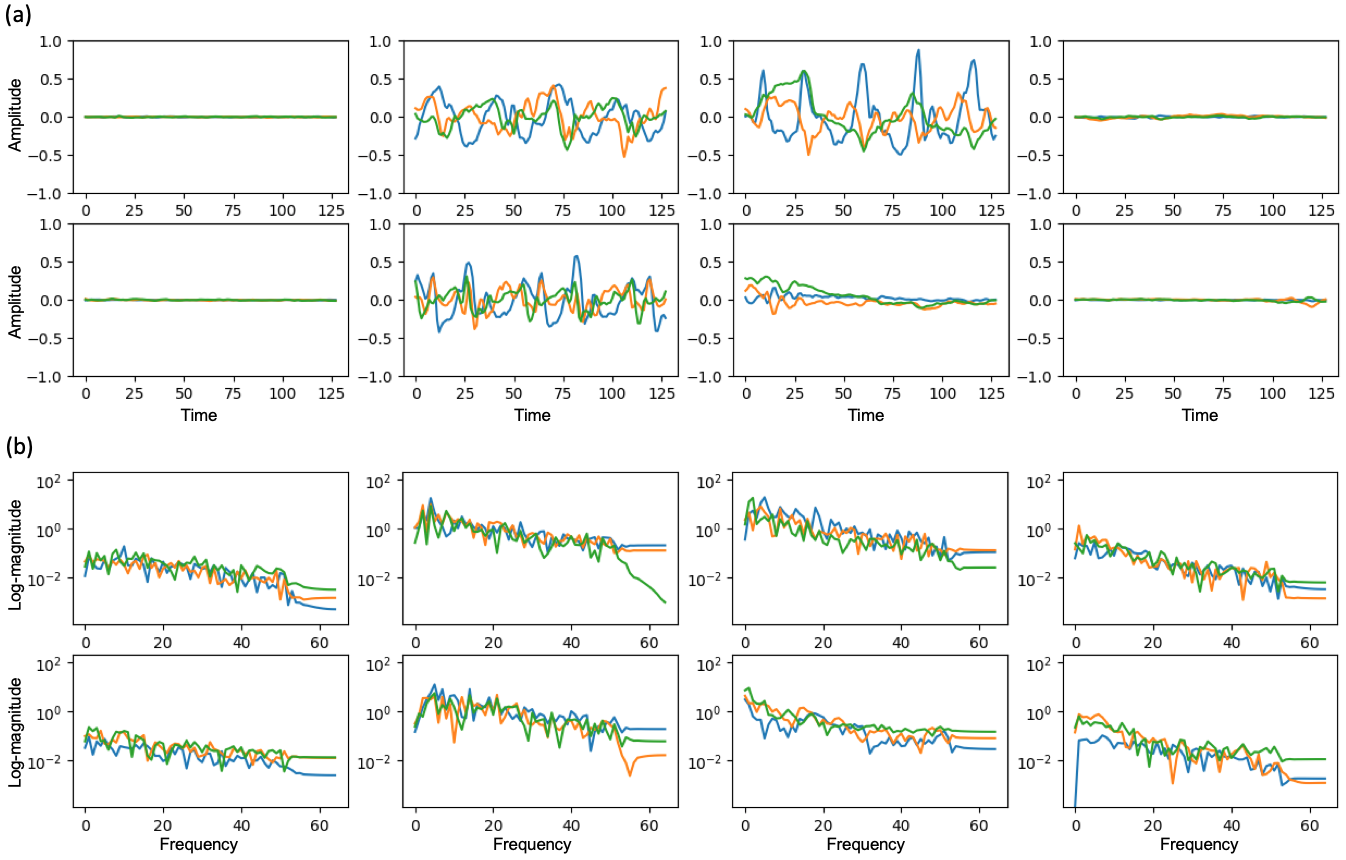}
  \caption{
  Visualization of eight representative examples from the HAR dataset in (a) the time and (b) frequency domain, without any data augmentation. Each color denotes a channel, and three channels are shown.
  }
  \label{fig:appendix_org}
\end{figure*}

\begin{figure*}[h]
  \centering
  \includegraphics[width=\textwidth]{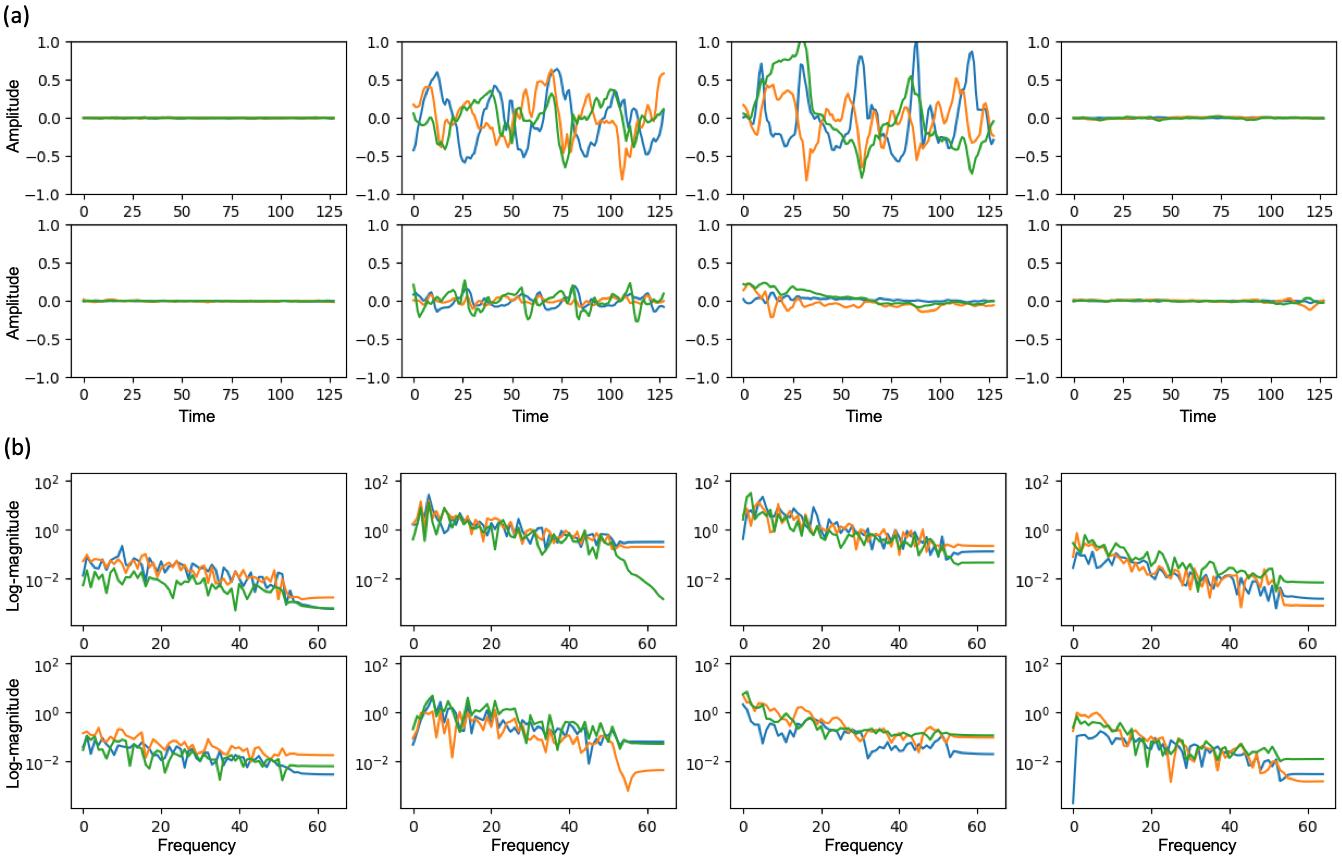}
  \caption{
  Visualization of eight representative examples from the HAR dataset in (a) the time and (b) frequency domain, augmented by Scaling. Each color denotes a channel, and three channels are shown.
  }
  \label{fig:appendix_scale}
\end{figure*}

\begin{figure*}[h]
  \centering
  \includegraphics[width=\textwidth]{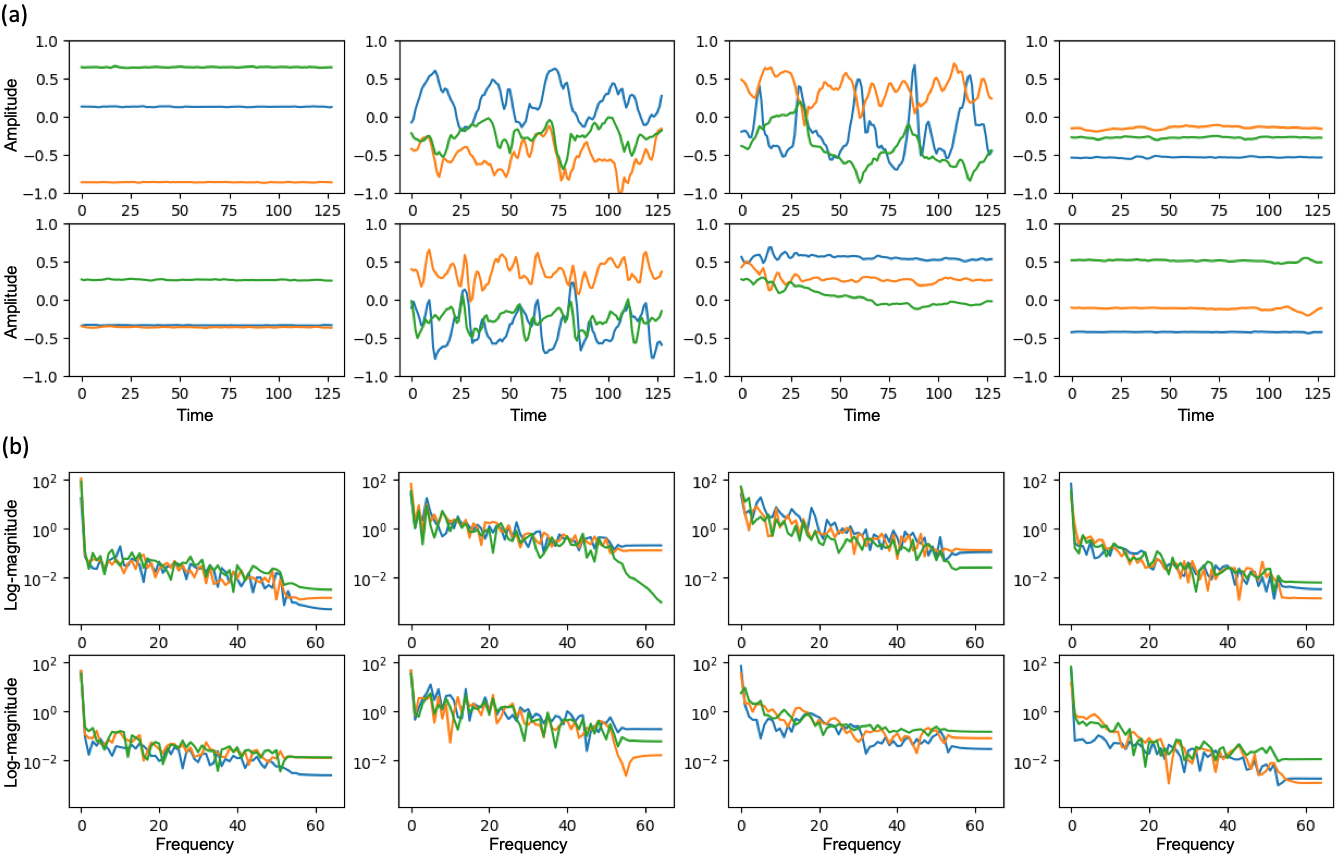}
  \caption{
  Visualization of eight representative examples from the HAR dataset in (a) the time and (b) frequency domain, augmented by Shifting. Each color denotes a channel, and three channels are shown.
  }
  \label{fig:appendix_shift}
\end{figure*}

\begin{figure*}[h]
  \centering
  \includegraphics[width=\textwidth]{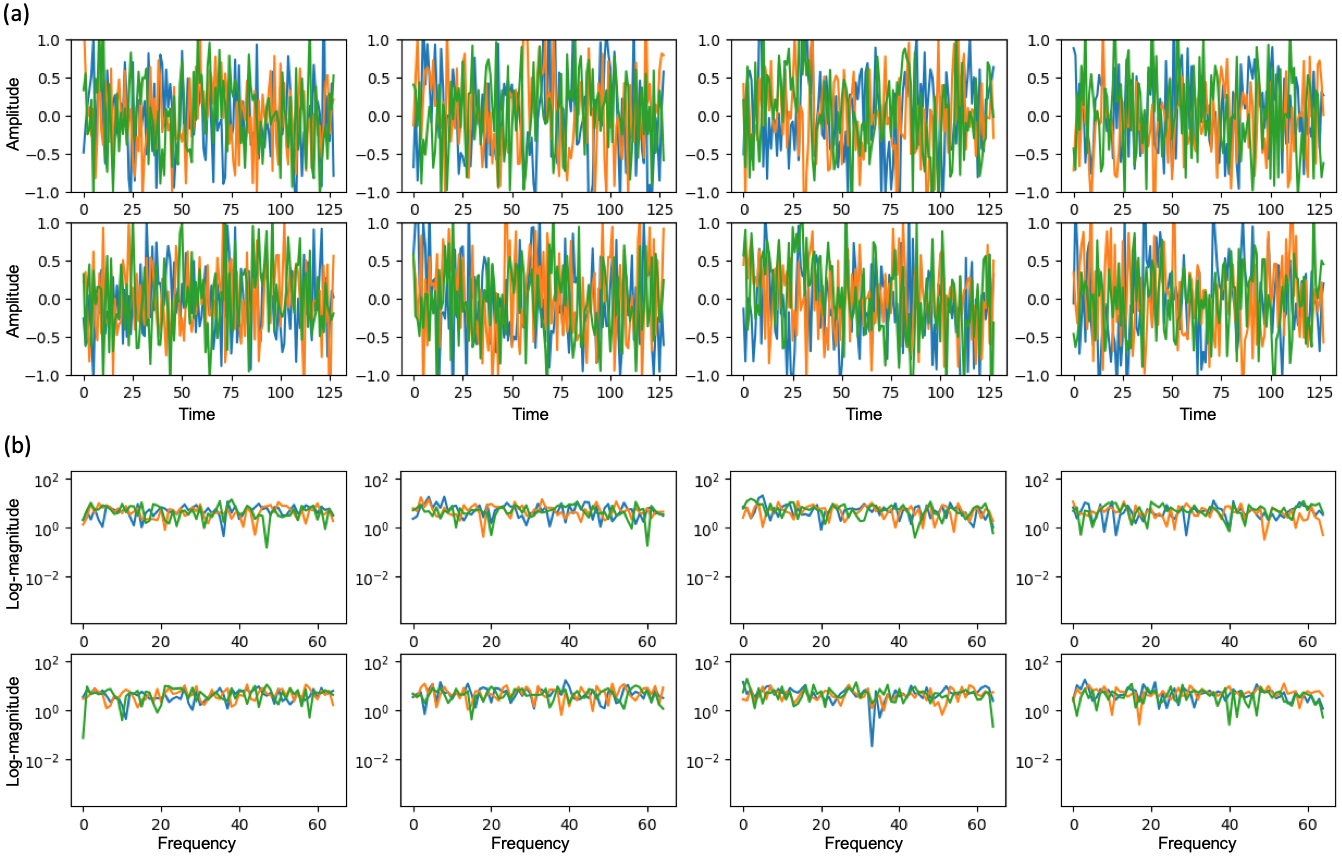}
  \caption{
  Visualization of eight representative examples from the HAR dataset in (a) the time and (b) frequency domain, augmented by Jittering. Each color denotes a channel, and three channels are shown.
  }
  \label{fig:appendix_j}
\end{figure*}

\begin{figure*}[h]
  \centering
  \includegraphics[width=\textwidth]{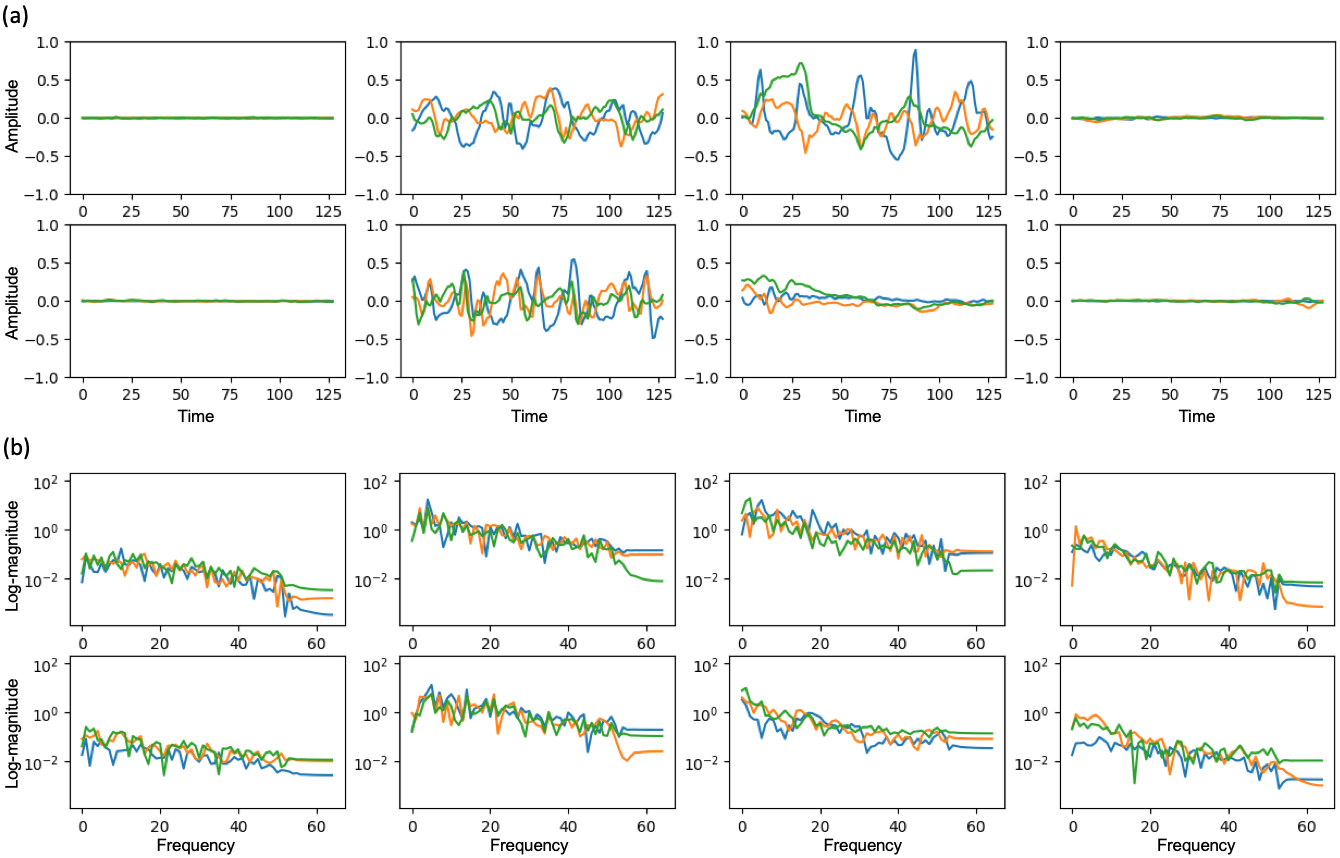}
  \caption{
  Visualization of eight representative examples from the HAR dataset in (a) the time and (b) frequency domain, augmented by Magnitude warping. Each color denotes a channel, and three channels are shown.
  }
  \label{fig:appendix_mw}
\end{figure*}

\begin{figure*}[h]
  \centering
  \includegraphics[width=\textwidth]{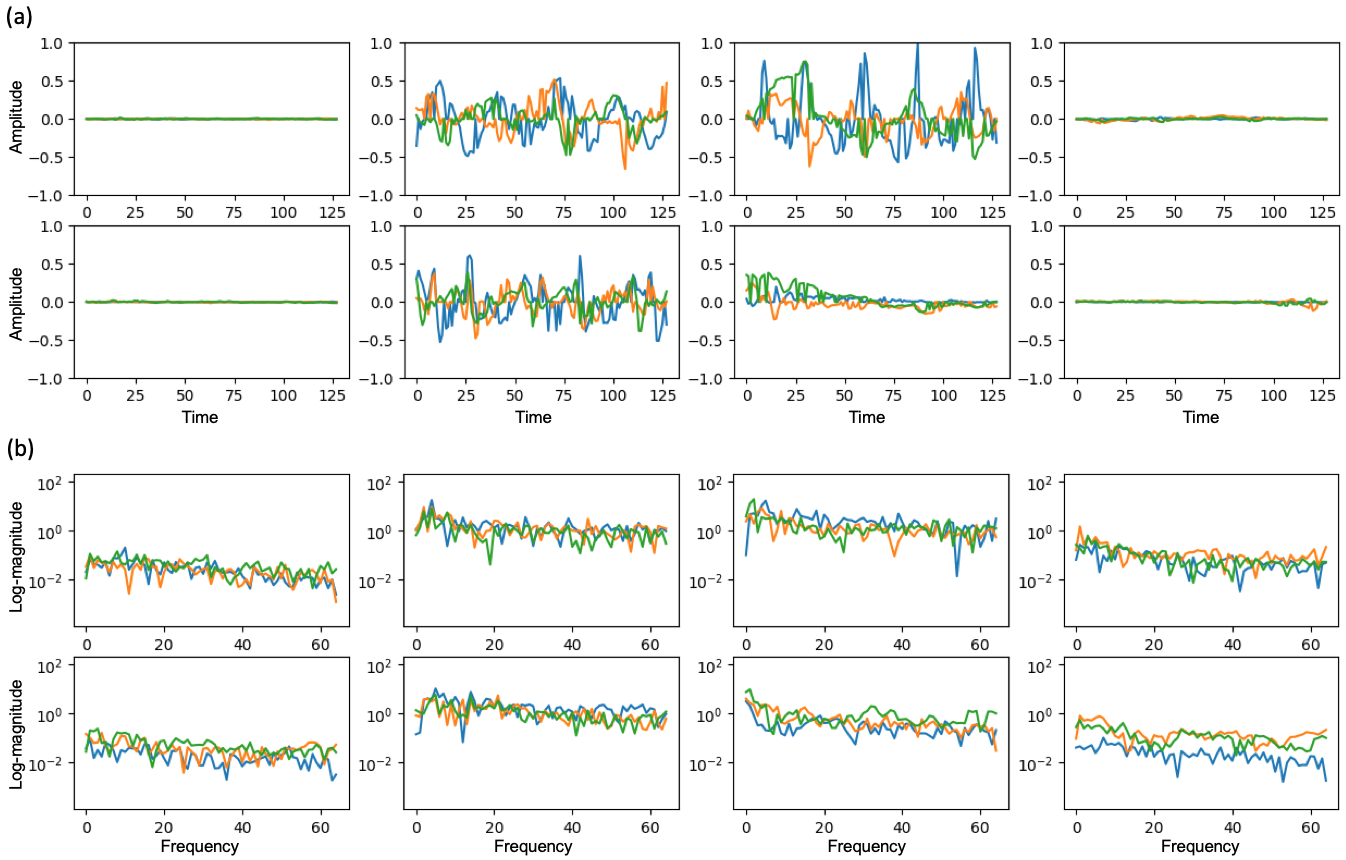}
  \caption{
  Visualization of eight representative examples from the HAR dataset in (a) the time and (b) frequency domain, augmented by Dropout. Each color denotes a channel, and three channels are shown.
  }
  \label{fig:appendix_td}
\end{figure*}

\begin{figure*}[h]
  \centering
  \includegraphics[width=\textwidth]{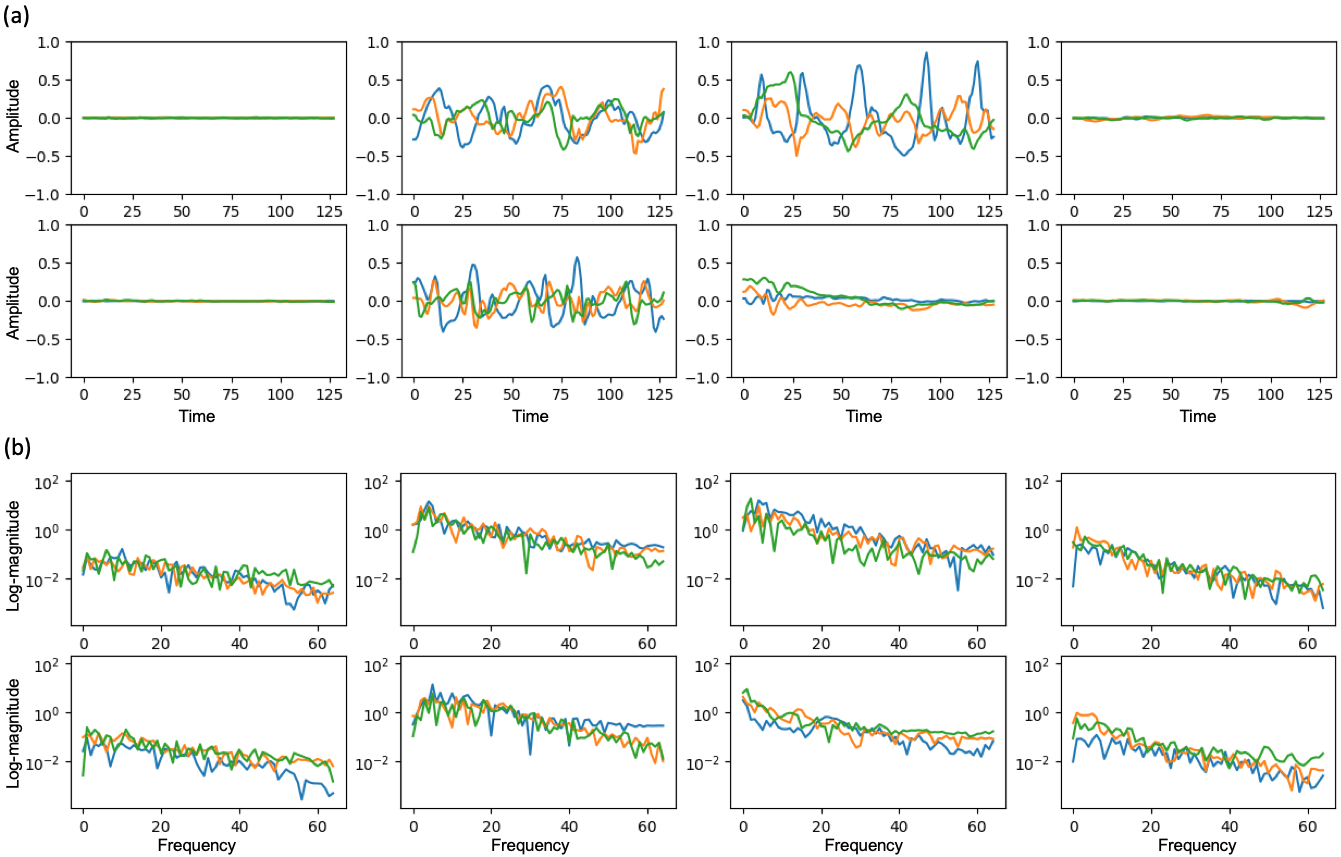}
  \caption{
  Visualization of eight representative examples from the HAR dataset in (a) the time and (b) frequency domain, augmented by Time warping. Each color denotes a channel, and three channels are shown.
  }
  \label{fig:appendix_tw}
\end{figure*}

\begin{figure*}[h]
  \centering
  \includegraphics[width=\textwidth]{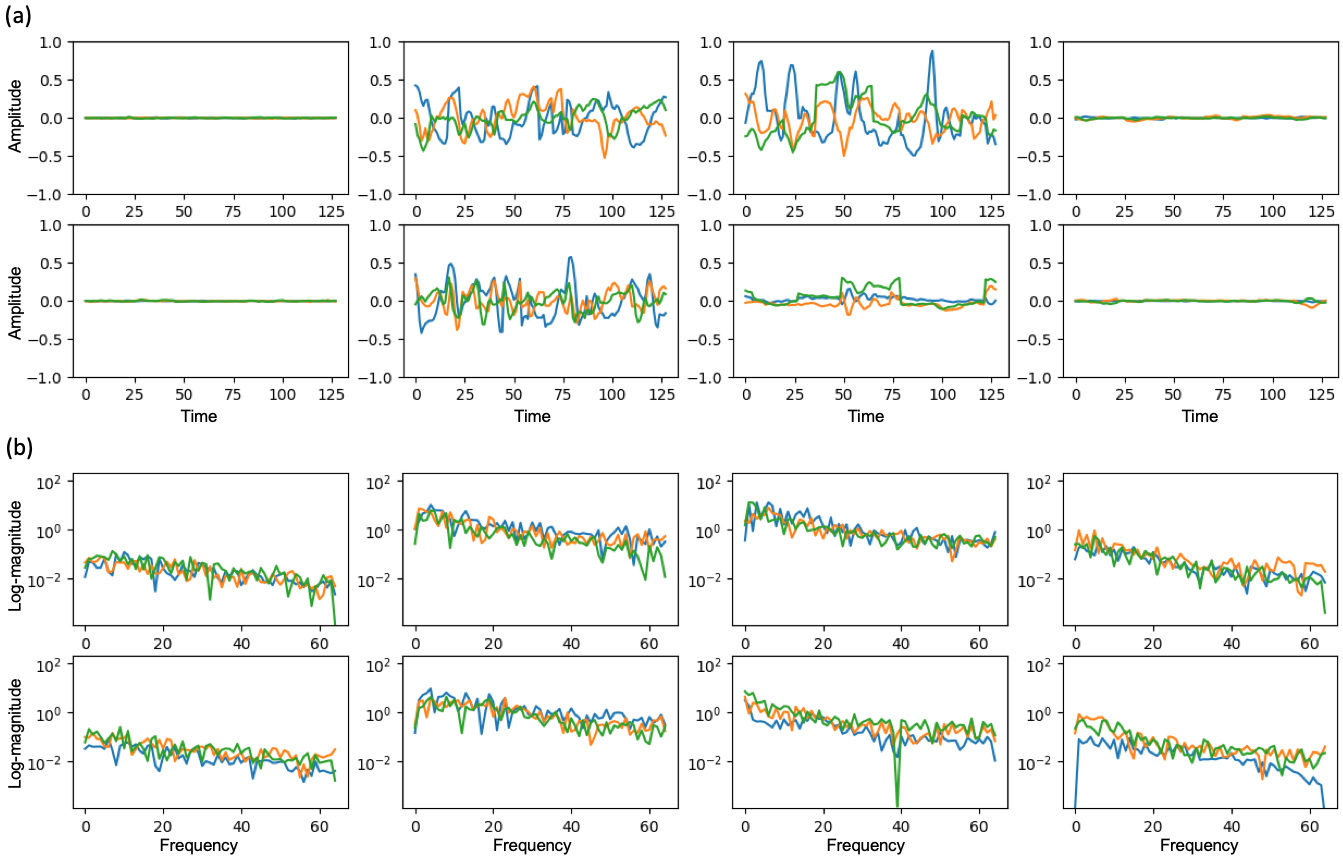}
  \caption{
  Visualization of eight representative examples from the HAR dataset in (a) the time and (b) frequency domain, augmented by Permutation. Each color denotes a channel, and three channels are shown.
  }
  \label{fig:appendix_p}
\end{figure*}

\begin{figure*}[h]
  \centering
  \includegraphics[width=\textwidth]{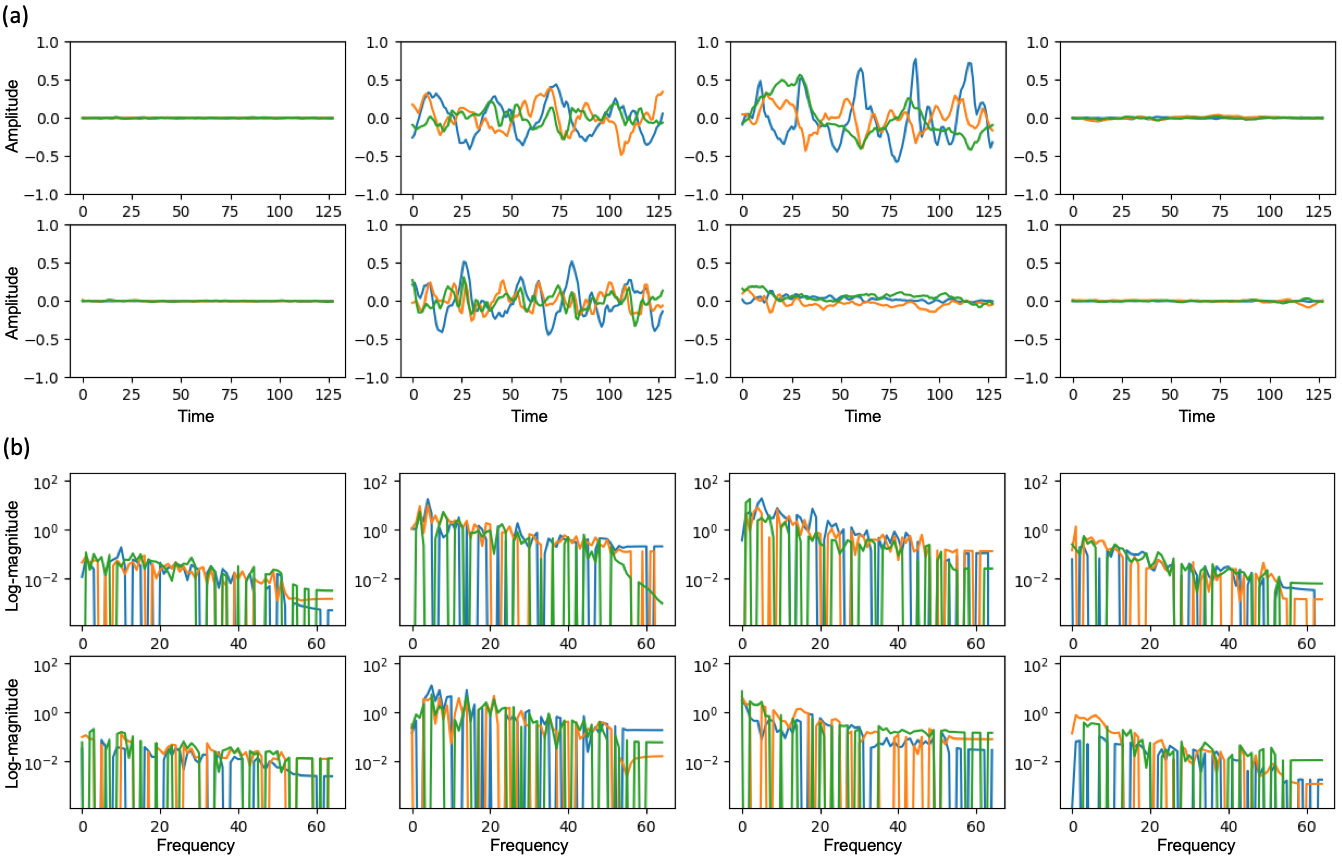}
  \caption{
  Visualization of eight representative examples from the HAR dataset in (a) the time and (b) frequency domain, augmented by Frequency masking. Each color denotes a channel, and three channels are shown.
  }
  \label{fig:appendix_fd}
\end{figure*}

We evaluated the effectiveness of the proposed \texttt{SimPSI} on seven random data augmentation techniques: Jittering \cite{um2017data}, Magnitude warping \cite{um2017data}, Dropout \cite{btsf}, Time warping \cite{um2017data}, Permutation \cite{um2017data}, Frequency masking \cite{fraug}, and composition of Scaling, Shifting, and Jittering \cite{cost}. We designed each technique to be applied or not by the probability $p = 0.5$ to incorporate original data in the training set. The following defines each technique and specifies parameter values. We notate an input time series as $x_t \in \mathbb{C}^{C \times L}$ and an augmented one as $x^{\prime}_t \in \mathbb{C}^{C \times L}$ where $C$ and $L$ denotes the number of channels and length of a time series.
Fig. \ref{fig:appendix_org} to \ref{fig:appendix_fd} visualize the original data and the corresponding augmented data with varying data augmentation techniques.

\textbf{Scaling.} An input time series is scaled by a random scalar $\epsilon$, sampled from a distribution $ N(1,0.5)$, where the augmented time series is $x^{\prime}_t = \epsilon x_t$.

\textbf{Shifting.} An input time series is shifted by a random scalar $\epsilon$, sampled from a distribution $ N(0,0.5)$, where the augmented time series is $x^{\prime}_t = x_t + \epsilon$.

\textbf{Jittering.} Gaussian noise $n_t \in \mathbb{R}^{C \times L}$, sampled from a distribution $N(0,0.5)$, is added to each time indices, where the augmented time series is $x^{\prime}_t = x_t + n_t$.

\textbf{Magnitude warping.} Random cubic polynomial $p_t\in \mathbb{R}^{C \times L}$ is elementwise multiplied with an input time series, where the augmented time series is $x^{\prime}_t = x_t \odot p_t$.

\textbf{Dropout.} An input time series is masked randomly by the probability $p = 0.2$.

\textbf{Time warping.} Random cubic polynomial $p_t\in \mathbb{R}^{C \times L}$ distorts the time interval of an input time series.

\textbf{Permutation.} An input time series is randomly partitioned and scrambled, where the maximum number of segments is $10$.

\textbf{Frequency masking.} An input time series is first transformed into the frequency domain. An input spectrum is masked randomly by the probability $p = 0.2$, then transformed back to the time domain.

\section{C. Model Architectures}
We chose three baseline models, 3-layer CNN, 2-layer LSTM, and 2-layer Transformer, to assess the effectiveness of \texttt{SimPSI} on different model architectures for the Human Activity Recognition task. A design of the CNN model adhered to \cite{tstcc}, so we provide a detailed configuration of the LSTM and Transformer model.
The LSTM model has two layers with a hidden dimension of 100, followed by a single linear layer for classification. The Transformer model has two transformer encoder layers, where each layer consists of a number of heads 2, dimension of feedforward network 256, and dropout with probability $p = 0.1$. It is followed by a single linear layer for classification purposes.

\clearpage

\clearpage

\section{D. Additional Results}

\subsection{D.1 Signal Demodulation}
We provided additional results for performance enhancement through \texttt{SimPSI} on the Simulation dataset (Table \ref{table:simulation}). The accuracy of Dropout is enhanced by 0.9\% using the spectrum-preservative map, while the random preservation map enhances it by 0.3\%. The accuracy of Time warping is increased by 0.8\% using the spectrum-preservative map, while the random preservation map decreases it by 0.7\%.
On the other hand, Permutation is not an appropriate augmentation technique for signal demodulation task that requires sequential decoding of the modulated signal, which significantly degrades the accuracy by 12.1\%. Surprisingly, the accuracy of Permutation is improved by 1.4\% using the spectrum-preservative map, while the random preservation map decreases it by 1.9\%. It highlights the potential of \texttt{SimPSI} as a simple yet strong strategy to improve the efficacy of any time series data augmentation technique.
The magnitude spectrum and saliency map show either marginal improvement or performance drop.

\begin{table}[h]
  \caption{Performance on Signal Demodulation (Simulation test set) using different random augmentations with and without \texttt{SimPSI}. Accuracy and its increment from not using augmentation are reported with three different seeds.}
  \label{table:simulation}
  \centering
  \begin{adjustbox}{width=\columnwidth}
  \begin{tabular}{lcc}
    \toprule
    Model & $Accuracy$ & $\Delta$ \\
    \midrule
    None                                           & 94.8 $\pm$ 0.1 & N/A \\
    \midrule
    Dropout \cite{btsf}                            & 93.4 $\pm$ 0.4 & \textminus1.4 \\
    + Random preservation map                      & 95.1 $\pm$ 0.5 & +0.3 \\
    + \texttt{SimPSI} (Magnitude spectrum)         & 94.9 $\pm$ 0.1 & +0.1 \\
    + \texttt{SimPSI} (Saliency map)               & 93.0 $\pm$ 0.4 & \textminus1.8 \\
    + \texttt{SimPSI} (Spectrum-preservative map)  & \textbf{95.7 $\pm$ 0.2} & \textbf{+0.9} \\
    \midrule
    Time warping \cite{um2017data}                 & 90.9 $\pm$ 1.0 & \textminus3.9 \\
    + Random preservation map                      & 94.1 $\pm$ 0.3 & \textminus0.7 \\
    + \texttt{SimPSI} (Magnitude spectrum)         & 94.9 $\pm$ 0.2 & +0.1 \\
    + \texttt{SimPSI} (Saliency map)               & 94.0 $\pm$ 0.1 & \textminus0.8 \\
    + \texttt{SimPSI} (Spectrum-preservative map)  & \textbf{95.6 $\pm$ 0.1} & \textbf{+0.8} \\
    \midrule
    Permutation \cite{um2017data}                  & 82.7 $\pm$ 2.9 & \textminus12.1 \\
    + Random preservation map                      & 92.9 $\pm$ 0.5 & \textminus1.9 \\
    + \texttt{SimPSI} (Magnitude spectrum)         & 94.8 $\pm$ 0.2 & 0 \\
    + \texttt{SimPSI} (Saliency map)               & 93.5 $\pm$ 0.6 & \textminus1.3 \\
    + \texttt{SimPSI} (Spectrum-preservative map)  & \textbf{96.2 $\pm$ 0.2} & \textbf{+1.4} \\
    \bottomrule
  \end{tabular}
  \end{adjustbox}
\end{table}

\subsection{D.2 Human Activity Recognition}

The performance improvement of data augmentations by \texttt{SimPSI} on the HAR dataset is summarized in Table \ref{table:har}. The accuracy of Time warping is enhanced by 1.0\% using the spectrum-preservative map, increased by 0.7\% using the magnitude spectrum and saliency map, while the random preservation map enhances it by 0.2\%. The accuracy of Permutation is improved by 0.2\% using the spectrum-preservative map, while the random preservation map decreases it by 0.8\%.

\begin{table}[h]
  \caption{Performance on Human Activity Recognition (HAR test set) using different random augmentations with and without \texttt{SimPSI}. Accuracy and its increment from not using augmentation are reported with three different seeds.}
  \label{table:har}
  \centering
  \begin{adjustbox}{width=\columnwidth}
  \begin{tabular}{lcc}
    \toprule
    Model & $Accuracy$ & $\Delta$ \\
    \midrule
    None                                           & 94.0 $\pm$ 0.8 & N/A \\
    \midrule
    Time warping \cite{um2017data}                 & 93.5 $\pm$ 0.5 & \textminus0.5 \\
    + Random preservation map                      & 94.2 $\pm$ 0.2 & +0.2 \\
    + \texttt{SimPSI} (Magnitude spectrum)         & 94.7 $\pm$ 0.9 & +0.7 \\
    + \texttt{SimPSI} (Saliency map)               & 94.7 $\pm$ 1.1 & +0.7 \\
    + \texttt{SimPSI} (Spectrum-preservative map)  & \textbf{95.0 $\pm$ 0.4} & \textbf{+1.0} \\
    \midrule
    Permutation \cite{um2017data}                  & 93.7 $\pm$ 0.6 & \textminus0.3 \\
    + Random preservation map                      & 93.2 $\pm$ 0.4 & \textminus0.8 \\
    + \texttt{SimPSI} (Magnitude spectrum)         & 93.6 $\pm$ 1.7 & \textminus0.4 \\
    + \texttt{SimPSI} (Saliency map)               & 93.4 $\pm$ 0.3 & \textminus0.6 \\
    + \texttt{SimPSI} (Spectrum-preservative map)  & \textbf{94.2 $\pm$ 0.4} & \textbf{+0.2} \\
    \bottomrule
  \end{tabular}
  \end{adjustbox}
\end{table}

\subsection{D.3 Sleep Stage Detection}
The performance increment of Dropout by \texttt{SimPSI} on the SleepEDF dataset is summarized in Table \ref{table:sleepedf}. The accuracy of Dropout is enhanced by 0.9\% using the spectrum-preservative map, while the random preservation map enhances it by 0.4\%.

\begin{table}[h]
  \caption{Performance on Sleep Stage Detection (SleepEDF test set) with and without \texttt{SimPSI}. Accuracy and its increment from not using augmentation are reported with three different seeds.}
  \label{table:sleepedf}
  \centering
  \begin{adjustbox}{width=\columnwidth}
  \begin{tabular}{lcc}
    \toprule
    Model & $Accuracy$ & $\Delta$ \\
    \midrule
    None                                           & 80.7 $\pm$ 0.1 & N/A \\
    \midrule
    Dropout \cite{btsf}                            & 80.8 $\pm$ 0.8 & +0.1 \\
    + Random preservation map                      & 81.1 $\pm$ 0.5 & +0.4 \\
    + \texttt{SimPSI} (Magnitude spectrum)         & 80.4 $\pm$ 0.9 & \textminus0.3 \\
    + \texttt{SimPSI} (Saliency map)               & 81.1 $\pm$ 0.8 & +0.4\\
    + \texttt{SimPSI} (Spectrum-preservative map)  & \textbf{81.6 $\pm$ 0.2} & \textbf{+0.9} \\
    \bottomrule
  \end{tabular}
  \end{adjustbox}
\end{table}

\subsection{D.4 Atrial Fibrillation Classification}
The performance enhancement of data augmentations by \texttt{SimPSI} on the Waveform dataset is summarized in Table \ref{table:waveform}. The accuracy of Jittering is improved by 0.4\% using the saliency map, while the random preservation map shows the same accuracy as not using augmentation. For Magnitude warping, none of the preservation maps improve the classification accuracy. We leave this deficiency as a future work to be resolved, which might be incurred by the wrong \texttt{SimPSI} hyperparameters choice since we did not focus on carefully choosing those. The accuracy of a composition of Scaling, Shifting, and Jittering is enhanced by 0.6\% using the saliency map and 0.5\% using the spectrum-preservative map, while the random preservation map decreases it by 0.2\%.

\begin{table}[h]
  \caption{Performance on Atrial Fibrillation Classification (Waveform test set) using different random augmentations with and without \texttt{SimPSI}. Accuracy and its increment from not using augmentation are reported with three different seeds.}
  \label{table:waveform}
  \centering
  \begin{adjustbox}{width=\columnwidth}
  \begin{tabular}{lcc}
    \toprule
    Model & $Accuracy$ & $\Delta$ \\
    \midrule
    None                                           & 94.7 $\pm$ 0.2 & N/A \\
    \midrule
    Jittering \cite{um2017data}                    & 94.2 $\pm$ 0.1 & \textminus0.5 \\
    + Random preservation map                      & 94.7 $\pm$ 0.2 & 0 \\
    + \texttt{SimPSI} (Magnitude spectrum)         & 94.6 $\pm$ 0.4 & \textminus0.1 \\
    + \texttt{SimPSI} (Saliency map)               & \textbf{95.1 $\pm$ 0.1} & \textbf{+0.4} \\
    + \texttt{SimPSI} (Spectrum-preservative map)  & 94.7 $\pm$ 0.1 & 0 \\
    \midrule
    Magnitude warping \cite{um2017data}            & 94.4 $\pm$ 0.3 & \textminus0.3 \\
    + Random preservation map                      & 94.3 $\pm$ 0.3 & \textminus0.4 \\
    + \texttt{SimPSI} (Magnitude spectrum)         & 94.2 $\pm$ 0.3 & \textminus0.5 \\
    + \texttt{SimPSI} (Saliency map)               & 94.4 $\pm$ 0.5 & \textminus0.3 \\
    + \texttt{SimPSI} (Spectrum-preservative map)  & \textbf{94.7 $\pm$ 0.4} & \textbf{0} \\
    \midrule
    Scale-Shift-Jittering \cite{cost}              & 92.0 $\pm$ 1.9 & \textminus2.7 \\
    + Random preservation map                      & 94.5 $\pm$ 0.4 & \textminus0.2 \\
    + \texttt{SimPSI} (Magnitude spectrum)         & 94.9 $\pm$ 0.6 & +0.2 \\
    + \texttt{SimPSI} (Saliency map)               & \textbf{95.3 $\pm$ 0.1} & \textbf{+0.6} \\
    + \texttt{SimPSI} (Spectrum-preservative map)  & 95.2 $\pm$ 0.3 & +0.5 \\
    \bottomrule
  \end{tabular}
  \end{adjustbox}
\end{table}

\section{E. Training Time Cost}
In this section, we compared the training time costs of different preservation maps applied to a composition of Scaling, Shifting, and Jittering on the HAR and SleepEDF datasets (Table \ref{table:time}). The magnitude spectrum requires a similar training time to the random preservation map, while the saliency map requires more than two times the training time than the magnitude spectrum. The spectrum-preservative map partially alleviates the computational burden but is still costly compared to the magnitude spectrum. We leave this limitation to further reduce the training time as future work.

\begin{table}[h]
  \caption{
  Training time on Human Activity Recognition (HAR train set) and Sleep Stage Detection (SleepEDF train set) with and without \texttt{SimPSI}.
  The total training times (second) are reported with three different seeds.
  }
  \label{table:time}
  \centering
  \begin{adjustbox}{width=\columnwidth}
  \begin{tabular}{lcc}
    \toprule
    Model & HAR & SleepEDF \\
    \midrule
    None                                           & 57 & 85\\
    \midrule
    Scale-Shift-Jittering \cite{cost}              & 59 & 89 \\
    + Random preservation map                      & 70 & 159 \\
    + \texttt{SimPSI} (Magnitude spectrum)         & 74 & 134 \\
    + \texttt{SimPSI} (Saliency map)               & 245 & 391 \\
    + \texttt{SimPSI} (Spectrum-preservative map)  & 162 & 364 \\
    \bottomrule
  \end{tabular}
  \end{adjustbox}
\end{table}

\section{F. Data Domain Dependency}
\begin{figure}[h]
  \centering
  \includegraphics[width=\columnwidth]{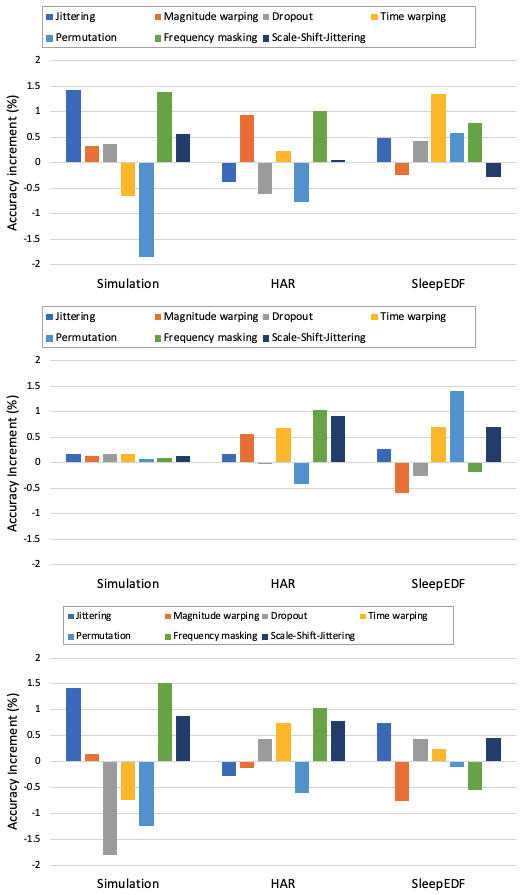}
  \caption{
  \texttt{SimPSI}'s dependency on data domain.
  Each plot shows the increment of classification accuracy of a baseline model after applying each data augmentation technique with different preservation maps.
  The top one corresponds to a random preservation map, the middle one corresponds to \texttt{SimPSI} (Magnitude spectrum), and the bottom one corresponds to \texttt{SimPSI} (Saliency map).
  }
  \label{fig:appendix_ddd}
\end{figure}

This section additionally provided data domain dependency of the magnitude spectrum and saliency map of \texttt{SimPSI}, as well as that of the random preservation map. The magnitude spectrum shows an overall improvement in the Simulation dataset, but the increment is marginal and does not apply to the HAR and SleepEDF datasets. The saliency map enhances performance by a large amount in some data augmentation techniques but highly depends on the data domain. The random preservation map also shows a high data domain dependency.
The results are summarized in Fig. \ref{fig:appendix_ddd}.

\end{document}